\documentclass[10pt]{IEEEtran}
\usepackage{tabularx}
\usepackage{booktabs}
\usepackage{graphicx}
\usepackage{graphics}
\usepackage{pbox}
\usepackage{dsfont}
\usepackage{psfrag}
\usepackage{color}
\usepackage{cite}

\usepackage[cmex10]{amsmath}
\usepackage{amssymb}
\usepackage{algorithmic}

\usepackage{array}

\usepackage{MnSymbol}

\usepackage{mdwmath}
\usepackage{mdwtab}

\usepackage{eqparbox}

\usepackage[tight,footnotesize]{subfigure}

\usepackage{stfloats}

\allowdisplaybreaks[4]

\begin{document}
%
\title{Cognitive Learning  of Statistical Primary Patterns  via Bayesian Network}

\author{Weijia~Han, Huiyan Sang, Min Sheng, Jiandong Li, and Shuguang Cui %
\thanks{Funding acknowledgement: The work was supported in part by DoD with grant HDTRA1-13-1-0029, by NSF with grants CNS-1343155, ECCS-1305979, and CNS-1265227,  by National Natural Science Foundation of China under Grants 61401320, 61231008, 91338114, 61172079, 61201141, 61301176, and 61328102, by the National High Technology Research and Development Program of China (863 Program) under Grant 2014AA01A701, by the 111 Project of China under Grant B08038.}
\thanks{W. Han, M. Sheng, and J. Li are with Broadband Wireless
Communications Lab. \& State Key Lab. (ISN), Information Science Institute, Xidian
University, Xi'an, Shaanxi, 710071, China (Emails: alfret@gmail.com, msheng@mail.xidian.edu.cn, jdli@pcn.xidian.edu.cn).}
\thanks{H. Sang is with the Department of Statistics, Texas A\&M University  (e-mail: huiyan@stat.tamu.edu).}
\thanks{S. Cui is with the Department of Electrical and Computer Engineering, Texas A\&M University (e-mail: cui@ece.tamu.edu). S. Cui is also a Distinguished Adjunct Professor at King Abdulaziz University in Saudi Arabia.}
}


\maketitle

\begin{abstract}
In cognitive radio (CR) technology, the trend of sensing is no longer to only detect the presence of active primary users. A large number of  applications demand for more comprehensive knowledge on  primary  user behaviors in spatial, temporal, and frequency domains. To satisfy  such requirements, we study the statistical relationship among primary users by introducing a Bayesian network (BN) based  framework. How to learn  such a BN structure  is a long standing issue, not fully understood  even in  the statistical  learning community. Besides, another key problem in this learning scenario  is that the CR has to identify how many variables are in the BN, which is usually considered as prior knowledge in statistical  learning applications. To solve such two issues simultaneously,  this paper  proposes a BN structure learning scheme  consisting of an efficient structure learning algorithm and a blind variable identification scheme. The proposed approach  incurs significantly  lower computational complexity compared with  previous ones, and is capable of determining the structure without assuming much prior knowledge about  variables. With this  result, cognitive users could efficiently understand  the statistical pattern of primary networks, such that more efficient cognitive protocols could be designed across different network layers.
\end{abstract}


%
\IEEEpeerreviewmaketitle

\newtheorem{proposition}{Proposition}
\newtheorem{theorem}{Theorem}
\newtheorem{lemma}{Lemma}
\newtheorem{corollary}{Corollary}

\section{Introduction}
Since the terminology was coined in 1999 \cite{788210}, cognitive radio (CR) has been developed for more than fifteen years, which has drawn   attention from  both academic and  industrial communities since it is intended to enable  smart use of the scarce  spectrum resource with the initial objective of maximizing spectrum utilization. Recently, cognitive network design goes beyond spectrum utilization and target at broader network objectives such as  higher quality of service, lower energy cost, etc. To achieve such new objectives, the statistical knowledge on the primary network status becomes necessary  \cite{6728604} for resource management and system control, which gets us closer to the ideal CR operation that integrates spectrum sensing, environment learning, statistical reasoning, and predictive acting. This will go beyond most of the existing CR sensing literature, which usually focus on detecting the presence of primary users only \cite{4840526, 6397631, 4453893}.

In practice, the network behavior is impacted by many factors which may change dynamically, such that the uncertainty of a network cannot be priorly represented by a certain statistical distribution. To cope with such an  issue, the statistical machine learning methodology becomes a feasible solution for understanding the network activity pattern. In this paper, we introduce the Bayesian network (BN) \cite{murphy1999modelling} structure learning method to obtain the statistical primary  networking pattern, via observing the on/off status of primary base stations.
The  Bayesian model has been well known in the field of artificial intelligence (AI) \cite{Daly:2011:RLB:2139625.2139626}. When considering the probability and uncertainty,  BN is  a distinct   technique  for modeling the complex interaction among real world facts \cite{mihajlovic2001dynamic}. In particular, the BN structure learning is an effective new modeling tool in both spatial and temporal domains. However, the associated  computational complexity is  high since it needs to evaluate the dependence between  each pair of variables in a target system and compute the corresponding conditional probability table\footnote{In statistics, the conditional probability table is defined for a set of discrete random variables to quantify the marginal probability of a single variable with respect to the others \cite{murphy2012machine}.}  \cite{Daly:2011:RLB:2139625.2139626}, which becomes  the  major drawback in applying   BN structure learning. In this paper,  we focus on reducing the computational load for efficiently learning the statistical behavior patterns  of  primary users. 

 For BN structure learning, the related algorithms could be sorted into two categories.  One is to use heuristic searching  to construct a probable model and then evaluate it by a scoring function \cite{shaughnessy2005evaluating}. The structure with the highest score is preferred as the learning outcome. The score-based approach of learning BNs has been proven as a NP-hard problem \cite{Chickering:2004:LLB:1005332.1044703}. 
 The other one is to use conditional independence test to measure every possible dependence relationships  one by one, and then determine the structure based on the evaluated dependence.
 In \cite{Cheng:2002:LBN:570380.570382},  the authors show  a much more efficient approach to learn the ordered BN by using mutual information to check the dependence of any possible pairs of nodes.  However, 
 this approach cannot be adaptively  adjusted when the number of variable changes. 
To overcome the  drawbacks of the current learning methods, we propose a structure learning algorithm based on a completely connected graph and the conditional mutual information, which could efficiently learn both the structure and the corresponding conditional probability table. In particular,    each pair of variables in the BN is  defined with a generalized relationship where the independence case is unified  as the weakest  dependence case. As a result,   the BN network structure is completely connected.  Accordingly, the network structure becomes very regular,  such that the conditional mutual information based  learning could be formulated as a sequence of closed-form function evaluations.  
With this,  our learning algorithm not only  has the same computational overhead as that in \cite{Cheng:2002:LBN:570380.570382}, but can also dynamically adapt to different numbers of variables. Moreover, for further  reducing the computational complexity, we simplify  the conditional mutual information function and explore   the prior knowledge that 
 the CR sensing results are   binary.

Besides the complexity problem, there are some special  issues in learning  BN structure in the context of CR, due to the lack of collaboration from the primary user side. In conventional learning cases,  many existing works assume  the number of  variables and their related observations are usually  prior  knowledge， and hence only focus on learning the structure among the variables.
 However, in CR, the observations are usually  collected without recording the correct  time epoch, causing the observation to be unidentifiable on  which time  that it  belongs to.   
This is mainly due to the fact that  the statistical period of the BN structure, which reflects the temporal pattern, is not known \emph{a priori}.  In essence, the above issues belong to the unsupervised  classification  problem, with a new challenge  to consider the missing time period. 
To address  this, we propose  a blind variable identification algorithm, combining  with the proposed structure learning algorithm, to learn the period via the fast Fourier transformation (FFT).  In conclusion, the proposed structure    and  period learning algorithms constitute a complete BN structure learning scheme, by which CR could understand the statistical pattern of primary base stations in both spatial and temporal domains. 

The remainder of the paper is organized as follows. In Section II, the system model is presented in details. Section
III briefly introduces the relationship between the BN  method and the wireless communication problem  in CR. In Section IV,  an efficient  algorithm is proposed  for jointly learning the structure and the corresponding conditional probability table. In Section V, an algorithm is proposed to obtain the statistical  period of BN structure. In Section VI, the simulation and learning results are presented to validate the proposed scheme. Finally, we conclude the paper in Section VII.

\section{System Model}
Our system model  consists of a primary cellular  network and multiple secondary sensors over an observation  area, of which the detailed specifications are given as follows.  

\begin{figure}[t]
  \centering
\includegraphics[width=6.5cm]{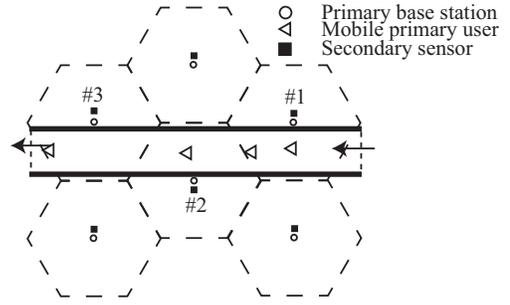}
  \caption{System Diagram}
 \label{f1}
\end{figure}

The primary  cellular network consists  of several  primary base stations and  multiple mobile primary users. The observation  area consists of  the cells  and a road as illustrated in Fig. \ref{f1}, where we  consider the mobile primary users only moving along a one-way (from right to left)  road for our case of  study.
The arrival of the users at the entrance of the road follows a Poisson process. The arrived mobile users pass along the road at a speed generated by a uniform distribution.  Once the mobile users move out the road,  they are no longer observed. Additionally, the primary network obeys  the following setup: 1) The primary base stations are located based on a pre-designed network deployment plan  (i.e., at the centers of the cells); 2) the primary users share a single channel and  access a primary base station depending on which cell they are located geographically;  3)  an ideal TDMA-based multiple access (MAC) scheme  is adopted, and the arrival and departure of primary data traffic at each user  follow a Poisson process and an exponential service law, respectively. Hence, the on/off status of a primary base station is determined by the overall  data traffic  generated from  the mobile primary users in its cell (here we only consider the uplink transmissions).   
 
A cognitive  secondary sensor is installed very  close to each primary base station, which implies that   the accuracy of sensing  could be assumed  perfect, such that no sensing errors are considered in this paper. The secondary sensors sense the on/off status of primary base stations periodically in a synchronous fashion.  
 Let set $\mathbb{M}:=\{1, 2, \cdots, M\}$ denote the observed primary base stations, and set $\mathbb{T}:=\{1, 2, \cdots, T\}$ denote the  sequence of time epoch $t$, $t\in{\mathbb{T}}$. In addition, let $f_{i, t}$ be a variable denoting the state of the $i$-th primary base station at time $t$, and $f_{i, t}\in \mathbb{O}:=\{0, 1\}$, where $0$ and $1$ represent the off and on statuses  respectively.  

\section{Cognitive Bayesian Network}
In CR, as we argued before, it is valuable to know the statistical behavior of the primary network. By exploring such  knowledge, the secondary network could exploit  the idle spectrum resource more efficiently and more broadly.
In our setup, each  secondary sensor could  obtain  a number of observations (or samples) about the on/off status of the observed primary base station, which is the key element of the primary network.  The existing  works with respect to CR sensing have been mainly focused  on the busy/idle status of a particular spectrum hole by  observing  a specific  primary base station. Contrastingly, our objective is to learn the statistical pattern of the spatial and temporal behavior of multiple networked primary base stations by mining the obtained observations over the whole network across  different time epochs.
To achieve our objective,   BN structure learning is deployed as the key methodology. The BN framework  has been known in the field of artificial intelligence and exploited in different expert systems to model complex interactions among causes and consequences, while BN structure learning aims to derive and quantify  the complex interaction from data. 

\subsection{Bayesian Network Approach}
 With BN,
the spatial and temporal interactions among primary base stations are expressed by a directed graph $\mathbb G=(\mathbb V, \mathbb E)$, where $\mathbb V$ is a finite, nonempty set whose elements are called the nodes denoting the variable  $f_{i, t}$, and $\mathbb E$ is a set of directed lines called the edges connecting the pairs of distinct elements in $\mathbb V$. If there is a directed edge from $f_{i_1, t_1}$ to $f_{i_2, t_2}$ where $f_{i_1, t_1}, f_{i_2, t_2} \in \mathbb E$, it means that  $f_{i_2, t_2}$ is impacted by $f_{i_1, t_1}$ i.e., $f_{i_2, t_2}$ depends on  $f_{i_1, t_1}$. 
 In BN analysis, such a directional  relationship  is  generally expressed by a conditional probability $P(f_{i_2, t_2}|f_{i_1, t_1})$.  For
the graphical BN, it has a distinguishing feature that:  For an arbitrary  $f_{i, t} \in \mathbb V$, it is conditionally  independent of the set of all other indirectly  connected nodes  given the set of all directly connected nodes. 
Apparently, if we have the complete knowledge on  $(\mathbb V, \mathbb E)$ together with all the values of $P(f_{i_2, t_2}|f_{i_1, t_1})$, the statistical pattern of the network behavior is readily available. 

\subsection{BN Structure Learning  in CR}
As explained above, for quantifying  the  BN and the related graph from observations, we need to determine both the variables (nodes) and the dependence of each pair of variables (edge). In many applications, the number of variables is priorly  known. Hence,  most of learning  results mainly  focus on how to learn the edges efficiently. In our CR scenario, since the observations are collected from the deployed sensors, it is easy to identify $\mathbb M$ for the range of $i$ in $f_{i, t}$.  However, due to the randomness of the  primary user number,  speed, and  traffic, the temporal information about $\mathbb T$, which defines the range for  $t$ of $f_{i, t}$, is not directly known. Thus,  BN learning in CR not only needs to efficiently mine the relationship and interaction among the variables, but also has to identify the temporal scale $T$ of BN. Since  $T$ is not known, we cannot simply sort the observations into the corresponding nodes. Hence, the unknown time scale $T$ becomes a critical issue in the proposed  CR BN learning. In machine learning language, such an issue belongs to the unsupervised classification  problem, which is widely considered difficult.

  \begin{figure}[t]
  \centering
\includegraphics[width=8.7cm]{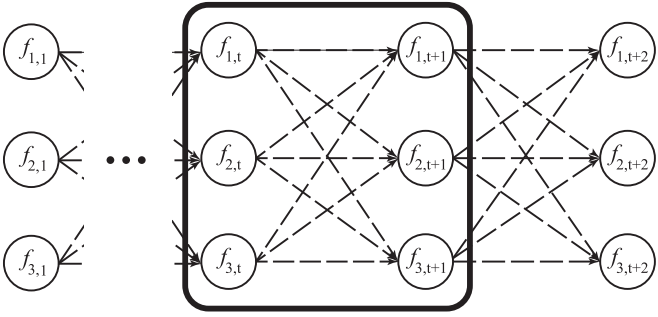}
  \caption{Learning CBN Structure, $M=3$. The dashed line means that the value  of an edge has not been determined.}
 \label{f3}
\end{figure}

For  edge  learning,  the traditional  learning methods are based on scoring  or dependence checking functions.  Given a scoring function, the computational complexity of determining  the BN structure increases exponentially when the number of variables increases. The score-based approach for learning Bayesian networks has been shown  NP-hard \cite{Chickering:2004:LLB:1005332.1044703}, which is a key challenge in the learning community. Recently,   a relatively  efficient way to learn an ordered BN is given in \cite{Cheng:2002:LBN:570380.570382} by using the conditional mutual information to check the dependence between  any possible pair of nodes. Formally, the conditional mutual information is defined as
\begin{equation}
\label{1.a.1}
\begin{aligned}
I(X; Y|Z) =& \sum_{z\in Z} \sum_{y\in Y} \sum_{x\in X}
      P_{X,Y,Z}(x,y,z) \\
      &\times\log \frac{P_Z(z)P_{X,Y,Z}(x,y,z)}{P_{X,Z}(x,z)P_{Y,Z}(y,z)},
\end{aligned}
\end{equation}
which is a measure of the mutual dependence between  variables $X$ and  $Y$ given variable $Z$, where the marginal, joint, and/or conditional probability mass functions are denoted by $P$ with  appropriate subscripts. Evidently, this method  demands for multiple nested for-loops 
for implementation, and the number of for-loops is determined by the number of variables in the BN. Hence, this method is  not  directly applicable  for the online learning case where the number of variables may be  time-varying.  Moreover, in \cite{Cheng:2002:LBN:570380.570382}, the conditional mutual information checking is performed for every  combination of all possible parent nodes of $X$. In other words, $I(f_{i,t}; f_{i,t-1}|\mathbb{F}_{t-1})$ is computed for all $\mathbb F_{t-1}$ where $\mathbb F_{t-1}$ denotes the $k$-combination of set $\{f_{1,t-1},\cdots, f_{M, t-1}\}$, $k\in\mathbb M$. The combination based checking  introduces  the huge computational overhead.
   
\subsection{Cognitive Bayesian Network Learning}  
Considering  the unknown $T$ and the high computational complexity  issues, we propose a BN model for our CR sensing case, and term it as cognitive BN (CBN). The CBN  has four characteristics: \textit{1})  It is a first-order BN and its nodes are ordered in the temporal domain; \textit{2}) its structure is completely  connected which means $\forall i\in\mathbb{M}, \forall t\in\mathbb{T}\backslash\{1\}$, there exists an edge between any $f_{i,t}$ and $f_{i, t-1}$; \textit{3}) the observation of  each variable is binary; \textit{4}) it has an unknown operation period. Here, ``ordered in the temporal domain" means that for a given $i$, $f_{i, t}, f_{i, t+1}\cdots$ are ordered in  $t, t+1\cdots$. Characteristics \textit{3} and \textit{4} are two direct outcomes from the system model. Thus we only explain characteristics \textit{1} and \textit{2}  below.

In our system model,  the user movement and data service behaviors of primary users at the current time epoch $t$ could be solely  determined by the system states in the former time epoch $t-1$. In other words, our system model  has the first-order Markov property,  which has been adopted previously \cite{bertsekas1987data}. In \cite{bertsekas1987data}, it is shown that a wireless communication network could be represented   by a Markov state transition system. 
Hence, we have  characteristic \textit{1} in our model, which leads  to  huge complexity reduction in checking the multiple ordered nodes. 
 Next, we explain characteristic \textit{2} in detail, which is unique and critical in further reducing the computational burden.

When learning the edges, the conventional approaches   usually consider the edges either existing or absent, by analyzing the directed dependence that is based on the empirical probabilities  generated from observations, where the edge weight could be  considered as a bi-level quantization of the directed dependence   \cite{6457478}.  
 In contrast, this paper first considers the existence of  every possible  edge in the first-order BN model, and then quantifies  the existence  with an analog value to reflect the dependence level between  any two nodes.   
In other words, we consider  independence as an extreme  case of dependence with edge value $=0$, which implies that   each pair of variables in the CBN has a generalized  relationship.
Based on such an approach, the CBN has a completely connected structure that is highly  regular, as shown in Fig. \ref{f3}. In the next section, we show that, by  exploring the regularity of the CBN structure, both the analog-valued edges and the conditional probability  table could be learned efficiently.

\section{Efficient Learning in CBN}
\label{IV}
In this section, we propose  an efficient learning algorithm  which could  correctly work under an arbitrary period $T$,  while the problem of an  unknown $T$ is studied in the next section. 
As explained before, high computational complexity is  a   critical issue in BN structure learning. It has been shown \cite{Cheng:2002:LBN:570380.570382} that the mutual information check based approach leads to the desired efficiency; but it cannot handle  a dynamic number of variables.  Here, we proposed an efficient learning algorithm  of the same complexity as the $I(X; Y|Z)$ based method, and can cope with the varying  number of variables.   

  When employing the conditional mutual information, every possible edge will be checked in turn. In other words, the number of possible edges affects  the learning overhead.   Recall that our CBN is a first-order BN and  ordered in the temporal domain. It means that, in CBN, the direction of edges is known and the edge only connects the adjacent nodes for a given $i$ in $f_{i, t}$ as shown in Fig. \ref{f3}. Hence, the computational complexity of learning such a CBN is proportional to learning a subgraph enclosed by the bold-line rectangle in Fig.  \ref{f3}. If we do not consider the direction of edges, this subgraph is called a clique  in a Markov network.  For the clarity of expression, we here  call the targeted  subgraph as a C-clique.  According to characteristic \textit{1}, the number of edges  in a CBN is $T$ times as that in a C-clique. Apparently, the computational  complexity  of learning  CBN is linearly proportional to the  overhead of learning a C-clique. Hence,  in the following study, we focus on how to  efficiently learn a C-clique.

\subsection{Learning a C-clique by Conventional Methods}
 \label{Conventional_Methods}
 
Before introducing our idea, we need to show how the current approaches  learn a CBN by using the conditional mutual information check, which is helpful for us to understand the computational complexity of learning a C-clique with the proposed  algorithm. {
In conventional  methods, when learning the edges in a C-clique, we need to determine the existence of possible edges one by one. For example, to check the edge between $f_{1,t}$ and  $f_{1,t+1}$ with a realization of  $M=3$  as shown in Fig. \ref{f3}, the corresponding conditional mutual information $I(f_{1,t}; f_{1, t+1}|\mathbb F_{t})$ is performed  by three times for $\mathbb F_{t}:=\{f_{2,t}\}$, $\mathbb F_{t}:=\{f_{3,t}\}$, and $\mathbb F_{t}:=\{f_{2,t},  f_{3,t}\}$. 

Actually, according to information theory,  there exists
 \begin{equation}
\label{1.a.2-1}
\begin{aligned}
I(f_{i,t}; f_{i, t+1}|\mathbb F_{t}) &=I(f_{i,t}; f_{i, t+1}|\mathbb F_{p, t}, \mathbb F_{t}\backslash \mathbb F_{p, t})\\
&=I(f_{i,t}; f_{i, t+1}|\mathbb F_{p, t})
\end{aligned}
\end{equation}
where $\mathbb F_{t}:=\{f_{1,t}, f_{2,t}, \cdots\}$ is a set containing all nodes at $t$, and $\mathbb F_{p,t}\subset \mathbb F_{t}$ is a set containing all parent nodes of $f_{i, t+1}$ at $t$. (\ref{1.a.2-1}) means that it is not necessary to check the mutual information conditioned on every possible $\mathbb F_{t}$.
To utilize such results, the proposed completely  connected structure shows a distinct merit that we only need perform the following checking once,
 \begin{equation}
\label{1.a.2}
\begin{aligned}
&I(f_{1,t}; f_{1, t+1}|f_{2,t}, f_{3,t}) \\
=&  \sum_{f_{1,t}} \sum_{f_{2,t}} \sum_{f_{3,t}} \sum_{f_{1,t+1}}
      P(f_{1,t}, f_{2,t}, f_{3,t}, f_{1,t+1}) \\
      &\times\log \frac{P(f_{2,t}, f_{3,t}) P(f_{1,t}, f_{2,t}, f_{3,t}, f_{1,t+1})}{P(f_{1,t}, f_{2,t}, f_{3,t}) P(f_{1,t+1}, f_{2,t}, f_{3,t})},
\end{aligned}
\end{equation}
where each probability is estimated by an empirical probability\footnote{For the clarity of expression, the empirical probability and the true probability are expressed by the same notation system.}. 

On the other hand, }the computational load of calculating  $I(f_{1,t}; f_{1, t+1}|f_{2,t}, f_{3,t})$ is mainly determined by using the related observations to  calculate  both  the empirical probability $P(f_{1,t}, f_{2,t}, f_{3,t}, f_{1,t+1})$ and the structure of $I(f_{1,t}; f_{1, t+1}|f_{2,t}, f_{3,t})$ function itself.
Apparently, the related  computation  requires  all the realizations of $\{f_{1,t}, f_{2,t}, f_{3,t} , f_{1, t+1}\}$, e.g. $\{f_{1,t}=0, f_{2,t}=0, f_{3,t}=0 , f_{1, t+1}=0\}$, $\{f_{1,t}=0, f_{2,t}=0, f_{3,t}=0 , f_{1, t+1}=1\}$, $\cdots$, $\{f_{1,t}=1, f_{2,t}=1, f_{3,t}=1, f_{1, t+1}=1\}$. As a result,  the computational complexity of measuring  an edge depends on computing $2^{(3+1)}$ times of $P(f_{1,t}, f_{2,t}, f_{3,t}, f_{1,t+1}) \log \frac{P(f_{2,t}, f_{3,t}) P(f_{1,t}, f_{2,t}, f_{3,t}, f_{1,t+1})}{P(f_{1,t}, f_{2,t}, f_{3,t}) P(f_{1,t+1}, f_{2,t}, f_{3,t})}$.\footnote{When implementing this calculation  in code, it needs $2^{(3+1)}$ for-loops.}
 In  general, since there are $M^2$ edges in a C-clique,  the computational complexity of learning  a C-clique is determined by running  $M^22^{(M+1)}$ times of $P(f_{1,t}, f_{2,t}, f_{3,t}, f_{1,t+1}) \log \frac{P(f_{2,t}, f_{3,t}) P(f_{1,t}, f_{2,t}, f_{3,t}, f_{1,t+1})}{P(f_{1,t}, f_{2,t}, f_{3,t}) P(f_{1,t+1}, f_{2,t}, f_{3,t})}$. Such outcome implies that the computational complexity of learning a C-clique could be reduced by improving the dependency checking function. 
 On the other hand, each term in (\ref{1.a.2}) is actually based on the conditional probability table considering that  the C-clique is completely connected.    
 When each term in (\ref{1.a.2}) is estimated by the empirical probabilities, the computational complexity of (\ref{1.a.2}) is proportional to that of computing  the conditional probability table.  Evidently, if we could efficiently   learn the complete conditional probability table and  quantify   each  edge  by  closed-form expressions and without nested for-loops,  the issues of varying variables and high computational complexity  will be solved and released, respectively. Based on the above two ideas, we next propose an efficient algorithm  
 based on analyzing the completely connected structure  and exploring the fact of binary observations. 
 
 \subsection{Efficient Learning Algorithm for  Conditional Probability Table}
 \label{efficient CPT} 
As explained before, the efficiency of the  mutual information based CBN learning is determined by two factors:  the measurement method of dependence and the adaptation to the number of variables. Among the two factors, the latter one plays a leading role.   In this subsection, we show our effort to handle the second  factor by utilizing   characteristic \textit{2} of our CBN.  

\subsubsection{Completely Connected  Structure and Its Benefit}
In a completely connected BN,  we need to compute $P(f_{i, t}|\text{pa}(f_{i, t}))$, $\forall i \in\mathbb{M}$ and  $\forall t \in\mathbb{{T}}$, to obtain the conditional probability table, where $\text{pa}(f_{i, t})$  means $f_{i, t}$'s parents that are the nodes \textbf{connecting to} $f_{i, t}$ directly. From a glance,  the completely connected BN is of high computational complexity because of the large number of edges. However, the fact is just the opposite. Since the structure of a completely connected BN is perfectly symmetry, the algorithm of computing the conditional probability table could  be designed efficiently, which will be discussed next.

 For a variable $f_{i, t}$, 
 consider its related observations  contained in a column vector   $\boldsymbol o_{i,t}$. For brevity, we take $M=2$ along with  binary observations to explain  our idea,  and then extend the related results to a general case with $M\ge 2$. From the perspective of frequentist probability,  the empirical   conditional probability table  of  $P(f_{i, t}|\text{pa}(f_{i, t}))$ in a C-clique is given by
 \begin{equation}
\label{2.a.1}
\begin{aligned}
P(f_{1, t}=1|f_{1, t-1}=1, f_{2, t-1}=1)=&\frac{\boldsymbol o_{1,t-1}^T\circ \boldsymbol o_{2,t-1}^T}{\boldsymbol o_{1,t-1}^T\boldsymbol o_{2,t-1}} \boldsymbol o_{1,t},
\end{aligned}
\end{equation}
where $\circ$ is the Hadamard product,
 \begin{equation}
\label{2.a.2}
\begin{aligned}
P(f_{1, t}=1|f_{1, t-1}=0, f_{2, t-1}=1)=&\frac{\bar{\boldsymbol o}_{1,t-1}^T\circ \boldsymbol o_{2,t-1}^T}{\bar{\boldsymbol o}_{1,t-1}^T\boldsymbol o_{2,t-1}} \boldsymbol o_{1,t},
\end{aligned}
\end{equation}
where $\bar{\boldsymbol o}_{1,t-1}^T=1-\boldsymbol o_{1,t-1}^T$,
 \begin{equation}
\label{2.a.3}
\begin{aligned}
P(f_{1, t}=1|f_{1, t-1}=1, f_{2, t-1}=0)=&\frac{\boldsymbol o_{1,t-1}^T\circ \bar{\boldsymbol o}_{2,t-1}^T}{\boldsymbol o_{1,t-1}^T\bar{\boldsymbol o}_{2,t-1}} \boldsymbol o_{1,t},
\end{aligned}
\end{equation}
where $\bar{\boldsymbol o}_{2,t-1}^T=1-\boldsymbol o_{2,t-1}^T$,
 \begin{equation}
\label{2.a.4}
\begin{aligned}
P(f_{1, t}=1|f_{1, t-1}=0, f_{2, t-1}=0)=&\frac{\bar{\boldsymbol o}_{1,t-1}^T\circ \bar{\boldsymbol o}_{2,t-1}^T}{\bar{\boldsymbol o}_{1,t-1}^T\bar{\boldsymbol o}_{2,t-1}} \boldsymbol o_{1,t},
\end{aligned}
\end{equation}
 \begin{equation}
\label{2.a.5}
\begin{aligned}
P(f_{2, t}=1|f_{1, t-1}=1, f_{2, t-1}=1)=&\frac{\boldsymbol o_{1,t-1}^T\circ \boldsymbol o_{2,t-1}^T}{\boldsymbol o_{1,t-1}^T\boldsymbol o_{2,t-1}} \boldsymbol o_{2,t},
\end{aligned}
\end{equation}
 \begin{equation}
\label{2.a.6}
\begin{aligned}
P(f_{2, t}=1|f_{1, t-1}=0, f_{2, t-1}=1)=&\frac{\bar{\boldsymbol o}_{1,t-1}^T\circ \boldsymbol o_{2,t-1}^T}{\bar{\boldsymbol o}_{1,t-1}^T\boldsymbol o_{2,t-1}} \boldsymbol o_{2,t},
\end{aligned}
\end{equation}
 \begin{equation}
\label{2.a.7}
\begin{aligned}
P(f_{2, t}=1|f_{1, t-1}=1, f_{2, t-1}=0)=&\frac{\boldsymbol o_{1,t-1}^T\circ \bar{\boldsymbol o}_{2,t-1}^T}{\boldsymbol o_{1,t-1}^T\bar{\boldsymbol o}_{2,t-1}} \boldsymbol o_{2,t},
\end{aligned}
\end{equation}
and
 \begin{equation}
\label{2.a.8}
\begin{aligned}
P(f_{2, t}=1|f_{1, t-1}=0, f_{2, t-1}=0)=&\frac{\bar{\boldsymbol o}_{1,t-1}^T\circ \bar{\boldsymbol o}_{2,t-1}^T}{\bar{\boldsymbol o}_{1,t-1}^T\bar{\boldsymbol o}_{2,t-1}} \boldsymbol o_{2,t}.
\end{aligned}
\end{equation}
 It is easy to see that the empirical  conditional probability could be calculated as multiplying   the observation $\boldsymbol o_{m,t}$  by a regular  arithmetic operator denoted by $\mathcal F$. We add an index $c$ to  $\mathcal F$ to express the condition, e.g., $\mathcal F_c$ with $c=00$ stands for  the  arithmetic operator under $f_{1, t-1}=0$ and $f_{2, t-1}=0$, such that  we have $\mathcal F_{00}=\frac{\bar{\boldsymbol o}_{1,t-1}^T\circ \bar{\boldsymbol o}_{2,t-1}^T}{\bar{\boldsymbol o}_{1,t-1}^T\bar{\boldsymbol o}_{2,t-1}}$.

From (\ref{2.a.1})-(\ref{2.a.4}) and (\ref{2.a.5})-(\ref{2.a.8}), we  see that  the  computation of the conditional probability table  is transformed to obtaining  the arithmetic operator $\mathcal F_c$ with every realization of index $c$, where the structure of $\mathcal F_c$ is very regular, which benefits from   characteristic \textit{2} of our CBN. Based on the regularity of $\mathcal F_c$, the edges no longer need to be learned one by one, which is explained   as follows.

\subsubsection{Binary Observation based Conditional Probability Table (BbCPT) Learning  Algorithm} In CR, the on/off behavior of a primary base station is   expressed by a binary value, which leads to our learning algorithm exploring this fact. 

Given  binary observations,  the number of realizations of $c$ is $2^M$, based on which we define 
 $x\in\{0, 1, \cdots, 2^M-1\}$. Accordingly,   for a given $x$, we could  use  (\ref{q3.1.2.2.0}) defined below  to generate a corresponding vector $\boldsymbol c_x$ leading to  $c=\boldsymbol c_x[1]\times \boldsymbol c_x[2]\times\cdots \times\boldsymbol c_x[M]$, with
 \begin{equation}
\label{q3.1.2.2.0}
\begin{aligned}
\boldsymbol c_x[i]=\left\lfloor \frac{x}{2^{i-1}}\right\rfloor\backslash 2
\end{aligned}
\end{equation}
where $\backslash$ denotes the modulo operation and $i\in\{1,2,\cdots, M\}$. 
For example, when $M=2$ and  $x=2$, $\boldsymbol c_2=[1 ~0]$, which  means $c=10$. Further, we could  generate a matrix  $\boldsymbol C$ where each row contains a realization of $c$   given $x$:
 \begin{equation}
\label{q3.1.2.2.1}
\begin{aligned}
\boldsymbol C[x, M-i+1]=\left\lfloor \frac{x}{2^{i-1}}\right\rfloor\backslash 2, i\in\mathbb M
\end{aligned}.
\end{equation}

On the other hand, when the observation is binary and we take $M=2$ as in the previous subsection, the numerators of $\mathcal F_{11}$, $\mathcal F_{10}$, $\mathcal F_{01}$, and $\mathcal F_{00}$ could be reformulated  as
 \begin{equation}
\label{q3.1.2.2.2}
\begin{aligned}
&\boldsymbol o_{1,t-1}^T\circ \boldsymbol o_{2,t-1}^T\\
=&\left\lfloor \frac{\boldsymbol o_{1,t-1}^T+ \boldsymbol o_{2,t-1}^T}{2}\right\rfloor\\
=&\left\lfloor \frac{1\boldsymbol o_{1,t-1}^T+ 1\boldsymbol o_{2,t-1}^T+0(1-\boldsymbol o_{1,t-1}^T)+ 0(1-\boldsymbol o_{2,t-1}^T)}{2}\right\rfloor\\
=&\left\lfloor \frac{\boldsymbol c_3[\boldsymbol o_{1,t-1}^T, \boldsymbol o_{2,t-1}^T]+(1-\boldsymbol c_3)(1-[\boldsymbol o_{1,t-1}^T, \boldsymbol o_{2,t-1}^T])}{2}\right\rfloor,
\end{aligned}
\end{equation}
 \begin{equation}
\label{q3.1.2.2.3}
\begin{aligned}
&\boldsymbol o_{1,t-1}^T\circ \bar{\boldsymbol o}_{2,t-1}^T\\
=&\left\lfloor \frac{\boldsymbol o_{1,t-1}^T+(1-\boldsymbol o_{2,t-1}^T)}{2}\right\rfloor\\
=&\left\lfloor \frac{1\boldsymbol o_{1,t-1}^T+ 0\boldsymbol o_{2,t-1}^T+0(1-\boldsymbol o_{1,t-1}^T)+ 1(1-\boldsymbol o_{2,t-1}^T)}{2}\right\rfloor\\
=&\left\lfloor \frac{\boldsymbol c_2[\boldsymbol o_{1,t-1}^T, \boldsymbol o_{2,t-1}^T]+(1-\boldsymbol c_2)(1-[\boldsymbol o_{1,t-1}^T, \boldsymbol o_{2,t-1}^T])}{2}\right\rfloor,
\end{aligned}
\end{equation}
 \begin{equation}
\label{q3.1.2.2.4}
\begin{aligned}
&\bar{\boldsymbol o}_{1,t-1}^T\circ \boldsymbol o_{2,t-1}^T\\
=&\left\lfloor \frac{(1-\boldsymbol o_{1,t-1}^T)+\boldsymbol o_{2,t-1}^T}{2}\right\rfloor\\
=&\left\lfloor \frac{\boldsymbol c_1[\boldsymbol o_{1,t-1}^T, \boldsymbol o_{2,t-1}^T]+(1-\boldsymbol c_1)(1-[\boldsymbol o_{1,t-1}^T, \boldsymbol o_{2,t-1}^T])}{2}\right\rfloor,
\end{aligned}
\end{equation}
 \begin{equation}
\label{q3.1.2.2.5}
\begin{aligned}
&\bar{\boldsymbol o}_{1,t-1}^T\circ \bar{\boldsymbol o}_{2,t-1}^T\\
=&\left\lfloor \frac{(1-\boldsymbol o_{1,t-1}^T)+(1-\boldsymbol o_{2,t-1}^T)}{2}\right\rfloor\\
=&\left\lfloor \frac{\boldsymbol c_0[\boldsymbol o_{1,t-1}^T, \boldsymbol o_{2,t-1}^T]+(1-\boldsymbol c_0)(1-[\boldsymbol o_{1,t-1}^T, \boldsymbol o_{2,t-1}^T])}{2}\right\rfloor,
\end{aligned}
\end{equation}
respectively. Let $\text{nom}(x)$ and $\text{dnom}(x)$ respectively denote the numerator  and  denominator of $x$, and $\boldsymbol{\mathcal F}$ be a matrix    containing all $\mathcal F_c$'s. Then, we arrive at a simple form for each  realization of $\mathcal F_c$ as
 \begin{equation}
\label{q3.1.2.2.6}
\begin{aligned}
&\text{nom}(\boldsymbol{\mathcal F})\\
=&[\text{nom}(\mathcal F_{00}); \text{nom}(\mathcal F_{01}); \text{nom}(\mathcal F_{10}); \text{nom}(\mathcal F_{11})]\\
=&\left\lfloor \frac{\boldsymbol C\boldsymbol O_{t-1}+(1-\boldsymbol C)(1-\boldsymbol O_{t-1})}{2}\right\rfloor
\end{aligned}
\end{equation}
where 
 \begin{equation}
\label{q3.1.2.2.7}
\begin{aligned}
\boldsymbol O_{t-1}=[\boldsymbol o_{1,t-1}, \boldsymbol o_{2,t-1}]^T,
\end{aligned}
\end{equation}
 \begin{equation}
\label{q3.1.2.2.8}
\begin{aligned}
\boldsymbol C
=&\begin{bmatrix}
0    & 0      \\
0 & 1\\
1   & 0\\
1    & 1 \\
\end{bmatrix}
\end{aligned},
\end{equation}
with $\boldsymbol C$  obtained from (\ref{q3.1.2.2.1}).

On the other hand, since $\text{nom}(\mathcal F_c)$ is a matrix, we have  $\text{dnom}(\mathcal F_c)=\text{nom}(\mathcal F_c)  \boldsymbol  1_N$, where $\boldsymbol  1_N$  is a column  vector of $N$ ones. Thus, the denominator of $\boldsymbol{\mathcal F}$ is given by
\begin{equation}
\label{q3.1.2.2.9}
\begin{aligned}
\text{dnom}(\boldsymbol{\mathcal F})=\text{nom}(\boldsymbol{\mathcal F})\boldsymbol 1_N.
\end{aligned}
\end{equation}

Extending (\ref{q3.1.2.2.6}) to an arbitrary value of  $M$ yields a general arithmetic operator $\boldsymbol{\mathcal F}$,
\begin{equation}
\label{q3.1.2.2.10}
\begin{aligned}
\boldsymbol{\mathcal F}=&\text{nom}(\boldsymbol{\mathcal F})./(\text{nom}(\boldsymbol{\mathcal F})\boldsymbol 1_N\otimes \boldsymbol 1_{N}^T),\\
\end{aligned}
\end{equation}
where $\otimes$ denotes  the Kronecker product, and
\begin{equation}
\label{q3.1.2.2.11}
\begin{aligned}
\text{nom}(\boldsymbol{\mathcal F})=&\left\lfloor \frac{\boldsymbol C\boldsymbol O_{t-1}+(1-\boldsymbol C)(1-\boldsymbol O_{t-1})]}{M}\right\rfloor,
\end{aligned}
\end{equation}
 \begin{equation}
\label{q3.1.2.2.12}
\begin{aligned}
\boldsymbol O_{t-1}=[\boldsymbol o_{1,t-1}, \boldsymbol o_{2,t-1}, \cdots, \boldsymbol o_{M,t-1}]^T.
\end{aligned}
\end{equation}
Hence, the empirical  conditional probability table could be  calculated  as 
 \begin{equation}
\label{q3.1.2.2.13}
\begin{aligned}
\boldsymbol{B}_{t-1}&=[\boldsymbol{B}_{1, t-1}, \cdots, \boldsymbol{B}_{M,t-1}]
&=\boldsymbol{\mathcal F}\boldsymbol O_{t}^T,
\end{aligned}
\end{equation}
where $\boldsymbol{B}_{i, t-1}$ is a column vector with size $2^M$ containing every probability that $f_{i, t}=1$ holds conditioned on $c$.  

Consequently, the  BbCPT learning  algorithm is given by (\ref{q3.1.2.2.1}) and (\ref{q3.1.2.2.10})-(\ref{q3.1.2.2.13}).
When compared  with conventional methods, the proposed  BbCPT  algorithm effectively  reduces the  number of for-loops to just two for-loops (with matrix computation), and the overall empirical  conditional probability table could be obtained by a sequence of closed-form function evaluations  efficiently. 

At the beginning of this subsection, we stated that  the total computational overhead is affected  by two factors,  with the latter one already discussed in this subsection. Next, we study the first  one to show how to efficiently measure the dependence between any pair of two variables in a C-clique.

\subsection{Efficient Dependence Measure}
\label{CBN Structure}
Given  discrete random variables $X$ with support  $\mathcal X$ and $Y$ with support  $\mathcal Y$, the conditional entropy between  $X$ and $Y$ is given by
 \begin{equation}
\label{q3.1.2.3.1}
\begin{aligned}
H(Y|X)=& \sum_{x\in\mathcal X}P(x)H(Y|X=x)\\
=&-\sum_{x\in\mathcal X} P(x)\sum_{y\in\mathcal Y}P(y|x)\log P(y|x)\\
=&\sum_{x\in\mathcal X, y\in\mathcal Y}P(x,y)\log \frac {P(x)} {P(x,y)}.
\end{aligned}
\end{equation}
If  $Y$ is completely determined by  $X$, we have $H(Y|X)=0$; if $Y$ is independent of $X$, we have $H(Y|X)=H(Y)$. Hence,  $H(Y|X)\in[0, H(Y)]$ reflects the dependence of $Y$ on $X$. 
According to information theory,  $H(Y|X)+H(X|Y)$ is equivalent to $I(X;Y)$ when measuring the dependence between $X$ and $Y$\footnote{In probability theory and information theory, $H(Y|X)+H(X|Y)$ is a  measure between variables for evaluating the variation of information or shared information distance.}. 

For the CR case where $\mathcal X:=\mathbb{O}$ and $\mathcal Y:=\mathbb{O}$, the conditional entropy  is given by
 \begin{equation}
\label{q3.1.2.3.2}
\begin{aligned}
H(Y|X)
=&-\sum_{x\in\mathcal X} P(x)P(y=1|x)\log P(y=1|x)\\
&-\sum_{x\in\mathcal X} P(x)P(y=0|x)\log P(y=0|x),\\
\end{aligned}
\end{equation}
which is always smaller or equal to $H(Y)$ over all possible $P(x)$.  According to (\ref{q3.1.2.3.2}), there is   $H(Y|X)=H(Y)$, i.e., $Y$ independs  of $X$,  when the following  holds. 
\begin{equation}
\label{q3.1.2.3.3}
\begin{aligned}
&\log{P(y|x=0)}-\log{P(y|x=1)}=0,
\end{aligned}
\end{equation}
for $y=0$ or $1$.

Similarly, we could conclude the same observations for $H(X|Y)$.
Evidently, given any values of $P(x)$ and $P(y)$,   smaller  values of $|\log{P(y=0|x=0)}-\log{P(y=0|x=1)|}$ and $|\log{P(y=1|x=0)}-\log{P(y=1|x=1)}|$ imply   larger values of $H(Y|X)$ or $H(X|Y)$, which further implies weaker  dependence between $Y$ and $X$.  
Therefore, we arrive at the following dependence metric, where $D_p(y; x)$ is termed as the conditional probability based dependence (CPbD),
\begin{equation}
\label{q3.1.2.3.5}
\begin{aligned}
D_p(y; x)=&\sum_{y\in\mathbb{O}}|\log{P(y|x=0)}-\log{P(y|x=1)}|\\
\end{aligned}
\end{equation}
which is  symmetric with respect to $P(y=1| x=1)$ and $P(y=1| x=0)$. 
The proposed CPbD $D_p(y;x)$ is not affected  by $P(y)$ and $P(x)$, and the value range of $D_p(y;x)$ is $[0, +\infty)$. For multiple-variable cases, the CPbD between  $x$ and $y$ conditioned on $z$ is given by
\begin{equation}
\label{q3.1.2.3.6}
\begin{aligned}
&D_p(y; x| z)\\
=&\sum_{y, z\in\mathbb{O}}|\log{P(y|z, x=0)}-\log{P(y|z,x=1)}|.
\end{aligned}
\end{equation}

Taking the example of $M=2$ in CBN, the  CPbD  between  $f_{1, t-1}$ and $f_{1, t}$ conditional on $f_{2, t-1}$ is given by
\begin{equation}
\label{q3.1.2.3.7}
\begin{aligned}
&D_p(f_{1, t}; f_{1, t-1}|f_{2, t-1})\\
=&|\log P(f_{1, t}=1|f_{1, t-1}=1, f_{2, t-1}=1)\\
 &-\log P(f_{1, t}=1|f_{1, t-1}=0, f_{2, t-1}=1)|\\
&+ |\log P(f_{1, t}=1|f_{1, t-1}=1, f_{2, t-1}=0)\\
 &~~~-\log P(f_{1, t}=1|f_{1, t-1}=0, f_{2, t-1}=0)|\\
&+ |\log P(f_{1, t}=0|f_{1, t-1}=1, f_{2, t-1}=1)\\
 &~~~-\log P(f_{1, t}=0|f_{1, t-1}=0, f_{2, t-1}=1)|\\
&+ |\log P(f_{1, t}=0|f_{1, t-1}=1, f_{2, t-1}=0)\\
 &~~~-\log P(f_{1, t}=0|f_{1, t-1}=0, f_{2, t-1}=0)|\\
\end{aligned}
\end{equation}
where each term inside the $\log$ operation could be directly obtained from the conditional probability table derived in Section \ref{efficient CPT}. Moreover,  each term  has a regular relative index $c$  corresponding to the conditional probability table.  As shown in the previous  subsection, the conditional probability table with respect to  $f_{1, t}$ is given as  $\boldsymbol B_{1, t-1}=[B_{00}, B_{01}, B_{10}, B_{11}]^T$. Let $\bar{\boldsymbol B}_{1, t-1}=1-{\boldsymbol B}_{1, t-1}$;  we have  $\log\boldsymbol B_{1, t-1}=[\log B_{00}, \log B_{01}, \log B_{10}, \log B_{11}]^T$ and $\log \bar{\boldsymbol B}_{1, t-1}={\log (1-\boldsymbol B}_{1, t-1})$. Then $D_p(f_{1, t};f_{1, t-1}|f_{2, t-1})$ and $D_p(f_{1, t};f_{2, t-1}|f_{1, t-1})$ can be calculated  as
\begin{equation}
\label{q3.1.2.3.8}
\begin{aligned}
&D_p(f_{1, t};f_{1, t-1}|f_{2, t-1})=(\left|\log\boldsymbol B_{1, t-1}^T\boldsymbol L_{1}\right| + \left|\log\bar{\boldsymbol B}_{1, t-1}^T\boldsymbol L_{1}\right| )\boldsymbol 1_{2},
\end{aligned}
\end{equation}
\begin{equation}
\label{q3.1.2.3.9}
\begin{aligned}
&D_p(f_{1, t};f_{2, t-1}|f_{1, t-1})=(\left|\log\boldsymbol B_{1, t-1}^T\boldsymbol L_{2}\right| + \left|\log\bar{\boldsymbol B}_{1, t-1}^T\boldsymbol L_{2}\right| )\boldsymbol 1_{2},
\end{aligned}
\end{equation}
where
\begin{equation}
\label{q3.1.2.3.10}
\begin{aligned}
\boldsymbol L_{1}=\begin{bmatrix}
0    & 1      \\
1    & 0 \\
0    & -1\\
-1   & 0 
\end{bmatrix},
\end{aligned}
\begin{aligned}
\boldsymbol L_{2}=\begin{bmatrix}
 0   &  1   \\
 0   &  -1  \\
 1   &  0   \\
-1   &  0 
\end{bmatrix},
\end{aligned}
\end{equation}

For every possible edge pointing to $f_{1, t}$, the corresponding  CPbD is given by 
\begin{equation}
\label{q3.1.2.3.12}
\begin{aligned}
&[D_p(f_{1, t};f_{1, t-1}|f_{2, t-1}), D_p(f_{1, t};f_{2, t-1}|f_{1, t-1})]\\
=&(\left|\log\boldsymbol B_{1, t-1}^T\boldsymbol L\right| + \left|\log\bar{\boldsymbol B}_{1, t-1}^T\boldsymbol L\right| )\boldsymbol S
\end{aligned}
\end{equation}
where
\begin{equation}
\label{q3.1.2.3.13}
\begin{aligned}
\boldsymbol L=[\boldsymbol L_{1}, \boldsymbol L_{2} ],
\end{aligned}
\end{equation}
\begin{equation}
\label{q3.1.2.3.14}
\begin{aligned}
\boldsymbol S=\boldsymbol I\otimes\boldsymbol 1_{2}.
\end{aligned}
\end{equation}

Similarly as the above,  it is easy to check  that $\boldsymbol L$ and $\boldsymbol S$ obtained for $f_{1, t}$ is still suitable for $f_{2, t}$. Thus, when $M=2$, the CPbD of the C-clique at $t-1$ is given by 
\begin{equation}
\label{q3.1.2.3.15}
\begin{aligned}
\boldsymbol D_{t-1}
=&(\left|\log\boldsymbol B_{t-1}^T\boldsymbol L\right| + \left|\log\bar{\boldsymbol B}_{t-1}^T\boldsymbol L\right| )\boldsymbol S
\end{aligned}
\end{equation}
where
\begin{equation}
\label{q3.1.2.3.16}
\begin{aligned}
\boldsymbol D_{t-1}=\begin{bmatrix}
 D_p(f_{1, t};f_{1, t-1}|f_{2, t-1})   &  D_p(f_{1, t};f_{2, t-1}|f_{1, t-1})   \\
 D_p(f_{2, t};f_{1, t-1}|f_{2, t-1})   &  D_p(f_{2, t};f_{2, t-1}|f_{1, t-1})  
\end{bmatrix},
\end{aligned}
\end{equation}
\begin{equation}
\label{q3.1.2.3.17}
\begin{aligned}
\boldsymbol B_{t-1}=[\boldsymbol B_{1, t-1}, \boldsymbol B_{2, t-1}],
\end{aligned}
\end{equation}
\begin{equation}
\label{q3.1.2.3.18}
\begin{aligned}
\boldsymbol L=[\boldsymbol L_{1}, \boldsymbol L_{2} ],
\end{aligned}
\end{equation}
\begin{equation}
\label{q3.1.2.3.19}
\begin{aligned}
\boldsymbol S=\boldsymbol I\otimes\boldsymbol 1_{2}.
\end{aligned}
\end{equation}

 In our system model, the behavior of each primary station follows a same birth-death process, which can be formulated by a Bayesian graph containing two nodes. It
means that if we only study the statistical pattern of one base station, the related Bayesian structure consists of one edges and two nodes which means $T=1$. Therefore, the value of $D_p(f_{1, t};f_{1, t-1}|f_{2, t-1})$ and $D_p(f_{2, t};f_{2, t-1}|f_{1, t-1})$ are same and keep constant over every C-clique of a whole BN graph which expresses the pattern of $M$ primary base stations. 
On the other hand,  the limited number of observations leads that the empirical $D_p(f_{1, t};f_{1, t-1}|f_{2, t-1})$ and $D_p(f_{2, t};f_{2, t-1}|f_{1, t-1})$ are different. Hence,   we normalize the entries in $\boldsymbol D_{t-1}$  normalized as follows for reflecting the spatial relationship clearly. 
 
\begin{equation}
\label{q3.1.2.3.19.1}
\begin{aligned}
\bar{\boldsymbol D}_{t-1}=&\Lambda^{-1}(\boldsymbol D_{t-1})\boldsymbol D_{t-1}\\
=&\begin{bmatrix}
 \frac{D_p(f_{1, t};f_{1, t-1}|f_{2, t-1})}{D_p(f_{1, t};f_{1, t-1}|f_{2, t-1})}   &  \frac{D_p(f_{1, t};f_{2, t-1}|f_{1, t-1})}{D_p(f_{1, t};f_{1, t-1}|f_{2, t-1})}    \\
 \frac{D_p(f_{2, t};f_{1, t-1}|f_{2, t-1})}{D_p(f_{2, t};f_{2, t-1}|f_{1, t-1}) }   &  \frac{D_p(f_{2, t};f_{2, t-1}|f_{1, t-1})}{D_p(f_{2, t};f_{2, t-1}|f_{1, t-1}) }  
\end{bmatrix},
\end{aligned}
\end{equation}
where $\Lambda(\boldsymbol D_{t-1})$ denotes the diagonal matrix with the same main diagonal as $\boldsymbol D_{t-1}$.

Consequently, for an arbitrary $M$, the normalized CPbD of C-clique at $t-1$ is given by
\begin{equation}
\label{q3.1.2.3.19.2}
\begin{aligned}
\bar{\boldsymbol D}_{t-1}=&\Lambda^{-1}(\boldsymbol D_{t-1})\boldsymbol D_{t-1},\\
\end{aligned}
\end{equation}
where
\begin{equation}
\label{q3.1.2.3.20}
\begin{aligned}
\boldsymbol D_{t-1}
=&(\left|\log\boldsymbol B_{t-1}^T\boldsymbol L\right| + \left|\log\bar{\boldsymbol B}_{t-1}^T\boldsymbol L\right| )\boldsymbol S,
\end{aligned}
\end{equation}
\begin{equation}
\label{q3.1.2.3.21}
\begin{aligned}
\boldsymbol B_{t-1}=[\boldsymbol B_{1, t-1}, \boldsymbol B_{2, t-1}, \cdots, , \boldsymbol B_{M, t-1}],
\end{aligned}
\end{equation}
\begin{equation}
\label{q3.1.2.3.22}
\begin{aligned}
\boldsymbol S=\boldsymbol I\otimes\boldsymbol 1_{M}.
\end{aligned}
\end{equation}

It is worth emphasizing  that  matrix $\boldsymbol L$ does not depend  on  the number of C-cliques, but only on the number of variables in one C-clique. Hence,   matrix $\boldsymbol L$  only needs to be computed once  for a given $M$. It could be computed offline prior to the  online CBN learning.  This paper does not study the optimal approach to obtain $\boldsymbol L$; alternatively, we provide  a  feasible method   to arrive at $\boldsymbol L$, given in Table \ref{L}.
 \begin{table}[h]
\centering
 \caption{Algorithm of Obtaining $\boldsymbol L$.}
  \label{L}
\begin{tabular}{ c | l }
  \hline                       
  1) & for $m=1:1:M$ \\
  2) &~~~$\boldsymbol e_1=\boldsymbol 0_{2^m\times 1} $, $\boldsymbol e_1[2^m, 1]=-1$, $\boldsymbol e_1[2^m/2,1]=1$,\\
  3) &~~~generate a circulant matrix   $\boldsymbol e$ based on $\boldsymbol e_1^T$,\\
  &~~~each row is a backward  shifting of $\boldsymbol e_1^T$, \\
  4) &~~~$\boldsymbol l_{1}=\boldsymbol e[1:m, :]^T$,\\
  5) &~~~for $n=2:1:m$\\
  6) &~~~~~~$\boldsymbol l_{n}=\boldsymbol I_{2\times 2}\otimes\boldsymbol L_{n-1}$, $\boldsymbol L_{n-1}=\boldsymbol l_{n-1}$,\\
  7) &~~~end\\
  8) &end\\
  9) &$\boldsymbol L=[\boldsymbol L_1, \boldsymbol L_2, \cdots, \boldsymbol L_M]$.\\
  \hline  
\end{tabular}
   \end{table}

  \subsection{Efficient C-clique Learning Procedure}
Based on the previously discussed BbCPT  and CPbD, the procedure of learning a C-clique is summarized as:  

$\\$
\begin{tabular}{ c | l }
  \hline 
   & Initialization of  $\boldsymbol C$,  $\boldsymbol L$, and $\boldsymbol S$; \\
  1) & Obtain the conditional probability table by    (\ref{q3.1.2.2.10})-(\ref{q3.1.2.2.13});\\ 
  2) & Obtain the CPbD of all edges by    (\ref{q3.1.2.3.19.2})-(\ref{q3.1.2.3.22}).\\         
  \hline  
\end{tabular}
$\\$

 As we have emphasized before,  $\boldsymbol C$ and  $\boldsymbol L$ are only required to calculate  once according to the number of variables in a C-clique, which means that they could  be  calculated and stored priorly.  When regarding $\boldsymbol C$ and $\boldsymbol L$ as prior knowledge,  the above procedure shows that our learning algorithm is just a sequence of closed-form function evaluations  and has a simpler dependence check function compared with that in \cite{Cheng:2002:LBN:570380.570382}. 
 Consequently, the proposed learning method has lower computational complexity and is capable of adapting to different numbers of variables. Note that the learned CPbD is a soft measure on the variable dependence, which will be useful in the next section to derive the general CBN learning algorithm. We could also now apply simple binary threshold over CPbD to decide the existence of the edges in the learned clique, as in \cite{Cheng:2002:LBN:570380.570382}.

 \section{Learning CBN with Unknown $T$}
\label{V}
The previous  section introduces  the CBN learning algorithm with a given $T$. Sometimes, the value of $T$ is not known \emph{a priori}. For example.   we do not know the exact value of $T$ in our CR system model. The uncertainty of  $T$ implies that  the number of variables in the temporal domain is unknown in the CBN model. Consequently, after a number of observations being collected,   we do not exactly  know  which temporal variables generated  the collected observations. 
In the   structure learning literature,  observations are usually assumed already associated with different variables correctly. As such,  the uncertainty of  $T$ is a challenging   issue, less studied in the BN community. For this problem,  we propose a heuristic but efficient solution as follows.   

To estimate  $T$, we first  formulate   this uncertainty issue into a mathematical problem.
Let $D_p(x)=\sum_{i, k\in{\mathcal M}}D_p(f_{i, t+x};f_{k, t}|f_{1:M, t}\backslash f_{k, t})$, which is the sum of the dependence measures over all edges between the variables at times $t$ and  $t+x$. In addition,  let $T_p$, $T_s$, and  $T^*$ denote the period of CPbD, the interval between two sequential observations of the same variable, and the minimal value of period  $T$, respectively. Note that all the periods in the set $\{T|T=kT^*, k\in\mathbb{Z}\}$ are  feasible for the BN structure (here $\mathbb{Z}$ denotes the positive integer set). But a longer period implies higher  computational complexity. 
Hence, we pick the minimal possible  value  $T^*$ as  the optimal choice. For $T^*$, it not only needs to ensure that  the observations are statistically periodic along the temporal variables   but also has to guarantee that the observations of a variable are statistically  independent over $T_s$.
Accordingly, when  $N\rightarrow \infty$, the optimal period $T^*$ could be obtained   by solving 
 \begin{equation}
\label{2.a.88}
\begin{aligned}
T^*&= \arg \min ~t \\
\text{st}.~~~~~~~~ t&=kT_p, \text{where} ~k\in \mathbb{Z},\\
t&\ge T_s^*-1,
\end{aligned}
\end{equation}
where $T_s^*=\min \{T_s|D_p(T_s)=0\}$. 

\subsection{Learning $T_s^*$}
 Specifically, the period $T$ should ensure that the observations of any variable $f_{i, t}$ are statistically  independent over $T_s$. Regarding  our system, the temporal correlation of two observations decreases as their interval increases. Obviously, it is better to set a long enough  period $T_s$ to ensure the statistically  independent condition. 
However, since the number of available observations $N$ is limited in practice,   we choose to estimate  $T_s$ and $T_p$ empirically. It is worth noting that: \textit{i}) it is impossible to deduce  the true probability or distribution of a variable from limited  observations, and \textit{ii}) the empirical statistical information is widely  used in real system analysis. From the perspective of both statistics and engineering,  the empirical probability is   valuable and useful  as long as it is close to the true one. In addition, it is impossible to find a feasible value of $T_s$ to ensure $D_p(T_s)=0$ using empirical probabilities. 
Hence, given the finite observations, we propose to obtain the empirical $T_s$ by finding the first valley value of the CPbD between $f_{1:M, t}$ and $f_{1:M, t+T_s}$, denoted by $\boldsymbol D_{t, t+T_s}$. Here   $f_{1:M, t}$ is a short form for $f_{1, t}, \cdots, f_{M, t}$. The  pseudo-code of the proposed algorithm denoted by $\mathcal{T}_s$  is given in Table \ref{ta1}, where $\frac{||\boldsymbol D_p||_1}{T_s}$ is the $\boldsymbol D_{t, t+T_s}$ value averaged  over $T_s$ and edges.
 \begin{table}[!h]
\centering
 \caption{Pseudo-code of $\mathcal{T}_s$.}
  \label{ta1}
\begin{tabular}{ c | l }
  \hline 
  1) & set $l=2$ \\
  2) & for $T_s=2:1:2^l+1$ \\
   &~~~for $t=1:1:T_s$ \\
   &~~~~~~$\boldsymbol D_p[t]= \boldsymbol D_{t, t+T_s}$;\\
   &~~~end\\
   &~~~$\boldsymbol D[T_s-1]=\frac{||\boldsymbol D_p||_1}{T_s}$;\\ 
   &~~~for $t=2:1:T_s-1$ \\
   &~~~~~~ if $\boldsymbol D_p[t]\le \boldsymbol D_p[t-1]$ and $\boldsymbol D_p[t]\le\boldsymbol D_p[t+1] $,\\
   &~~~~~~~then break and output $T_s^*=t$;\\
   &~~~end\\ 
   &end\\
3) &if $T_s=2^l+1$,\\
   &then set $l=l+1$ and goto 2).\\
  \hline  
\end{tabular}
   \end{table}

Next, we give our algorithm that obtains $T_p$ by exploring  the regularity of CPbD. 

\subsection{Learning $T_p$}
As discussed earlier, the CPbD  has the capability of reflecting the dependence between variables. Hence, when  finding the correct value of $T_p$, the CPbD should demonstrate  a certain regularity that could help with learning $T_p$.  This subsection first shows the existence of CPbD regularity  mathematically.
 We need to emphasize   that the observations are assumed to have certain  correlation in the user and temporal domains, though  we do not know the statistical characteristics of such correlation, since if all observations are independent, there will be no edges in the BN structure.
To ensure that the unknown $T_p$ could be obtained by learning from the observations, we have the following theoretical results. 

First, we derive \textit{Proposition \ref{tt2}} and  \textit{Proposition \ref{tt3}} to show that $P(f_{i, t+1}=1|f_{i, t}=1)$  is regular  when considering $T$ as a variable, which means that the calculation of the empirical  $P(f_{i, t+1}=1|f_{i, t}=1)$ has a regular pattern  either. Let $\mathbb{T}_f$ and $\mathbb{T}_r$ denote the sets containing the false and true periods respectively.  Obviously, there is $\mathbb{T}:=\mathbb{T}_f \cup\mathbb{T}_r$. And let $\mathbb Z_{e}:=\{2k: k~ \text{is positive integer}\}$ and $\mathbb Z_{o}:=\{2k+1: k~ \text{is positive integer}\}$. Additionally, consider $T_{f}'$ is a variable similar as $T_{f}$.
\begin{proposition}
\label{tt2}
When $T_{f},T_{f}'\in\mathbb{T}_f$,  $ T_{r}\in \mathbb{T}_r \cap\mathbb Z_{o}$,  there exist:
\begin{description}
\item[\textit{P1.1})]  $P(f_{i, t+1}=1|f_{i, t}=1)|_{T=T_{f}}\ne P(f_{i, t+1}=1|f_{i, t}=1)|_{T=T_{r}}$.
\item[\textit{P1.2})]  $P(f_{i, t+1}=1|f_{i, t}=1)|_{T=T_{f}}= P(f_{i, t+1}=1|f_{i, t}=1)|_{T=T_{f}'}$.
\end{description}
\end{proposition}
\begin{proposition}
\label{tt3}
When $T_{f},T_{f}'\in\mathbb{T}_f$,  $ T_{r}\in \mathbb{T}_r\cap\mathbb Z_{e}$,  there exist:
\begin{description}
\item[\textit{P2.1})]   $P(f_{i, t+1}=1|f_{i, t}=1)|_{T=T_{f}}\ne P(f_{i, t+1}=1|f_{i, t}=1)|_{T=T_{r}}$.
\item[\textit{P2.2})]   $P(f_{i, t+1}=1|f_{i, t}=1)|_{T=T_{f}}= P(f_{i, t+1}=1|f_{i, t}=1)|_{T=T_{f}'}$ under the condition that  $T_{f}, T_{f}'\in\mathbb Z_{e}$.
\item[\textit{P2.3})]   $P(f_{i, t+1}=1|f_{i, t}=1)|_{T=T_{f}}= P(f_{i, t+1}=1|f_{i, t}=1)|_{T=T_{f}'}$ under the condition that  $T_{f}, T_{f}'\in\mathbb Z_{o}$.
\item[\textit{P2.4})]   $P(f_{i, t+1}=1|f_{i, t}=1)|_{T=T_{f}}\ne P(f_{i, t+1}=1|f_{i, t}=1)|_{T=T_{f}'}$ under the condition that  $T_{f}\in\mathbb Z_{o}$ and $T_{f}'\in\mathbb Z_{e}$.
\end{description}
\end{proposition}

\begin{IEEEproof}
We   prove \textit{Proposition \ref{tt2}} at first.
Let $\{o_{1,1}, o_{1, 2}, \cdots, o_{1, N}\}$ denote the $N$  temporal observations of user-1. Then, given the period $T$,  we fold this time series at $T$. For example,
 \begin{itemize}
\item When $T=2$, we have $\boldsymbol o_{1,1}|_{T=2}=[o_{1,1}, o_{1,3}, o_{1,5},\cdots]^T$ and $\boldsymbol o_{1, 2}|_{T=2}=[o_{1,2}, o_{1, 4}, o_{1, 6},\cdots]^T$. 
\item When $T=3$, we have  $\boldsymbol o_{1,1}|_{T=3}=[o_{1,1}, o_{1, 4}, o_{1, 7},\cdots]^T$, $\boldsymbol o_{1, 2}|_{T=3}=[o_{1,2}, o_{1, 5}, o_{1, 8},\cdots]^T$, and $\boldsymbol o_{1, 3}|_{T=3}=[o_{1, 3}, o_{1, 6}, o_{1, 9},\cdots]^T$. 
\item When $T=4$, we have  $\boldsymbol o_{1,1}|_{T=4}=[o_{1,1}, o_{1, 5}, o_{1, 9},\cdots]^T$, $\boldsymbol o_{1, 2}|_{T=4}=[o_{1,2}, o_{1, 6}, o_{1, 10},\cdots]^T$, $\boldsymbol o_{1, 3}|_{T=4}=[o_{1, 3}, o_{1, 7}, o_{1, 11},\cdots]^T$, and  $\boldsymbol o_{1, 4}|_{T=4}=[o_{1, 4}, o_{1, 8}, o_{1, 12},\cdots]^T$.
\item When $T=5$, there exist  $\boldsymbol o_{1,1}|_{T=5}=[o_{1,1}, o_{1, 6}, o_{1, 11},\cdots]^T$, $\boldsymbol o_{1, 2}|_{T=5}=[o_{1,2}, o_{1, 7}, o_{1, 12},\cdots]^T$, $\boldsymbol o_{1, 3}|_{T=5}=[o_{1, 3}, o_{1, 8}, o_{1, 13},\cdots]^T$, $\boldsymbol o_{1, 4}|_{T=5}=[o_{1, 4}, o_{1, 9}, o_{1, 14},\cdots]^T$, and  $\boldsymbol o_{1, 5}|_{T=5}=[o_{1, 5}, o_{1, 10}, o_{1, 15},\cdots]^T$.
\item When $T=6$, there exist  $\boldsymbol o_{1,1}|_{T=6}=[o_{1,1}, o_{1, 7}, o_{1, 13},\cdots]^T$, $\boldsymbol o_{1, 2}|_{T=6}=[o_{1,2}, o_{1, 8}, o_{1, 14},\cdots]^T$, $\boldsymbol o_{1, 3}|_{T=6}=[o_{1, 3}, o_{1, 9}, o_{1, 15},\cdots]^T$, $\boldsymbol o_{1, 4}|_{T=6}=[o_{1, 4}, o_{1, 10}, o_{1, 16},\cdots]^T$, $\boldsymbol o_{1, 5}|_{T=6}=[o_{1, 5}, o_{1, 11}, o_{1, 17},\cdots]^T$, and  $\boldsymbol o_{1, 6}|_{T=6}=[o_{1, 6}, o_{1, 12}, o_{1, 18},\cdots]^T$.
\end{itemize}

Accordingly, when $T_{r}=5$, we have
\begin{equation}
\label{q4.1}
\begin{aligned}
&P(f_{1, 2}=1|f_{1, 1}=1, T_{f}=2, T_{r}=5)\\
=&\lim_{N\to \infty} \frac{\boldsymbol o_{1, 1}^T\boldsymbol o_{1, 2}}{\boldsymbol o_{1, 1}^T\boldsymbol 1}\bigg |_{T=T_{f}=2}\\
=&\lim_{N\to \infty}\frac{\sum_{t=1}^{4}\boldsymbol o_{1, t}^T\boldsymbol o_{1, t+1}+\boldsymbol o_{1, 5}^TS(\boldsymbol o_{1, 1})}{\left(\sum_{t=1}^{5}\boldsymbol o_{1, t}^T\right)\boldsymbol 1}\bigg |_{T=T_{r}=5},
\end{aligned}
\end{equation}
where $S(.)$ denotes an operation  shifting the entries of a vector with a single step  circularly,
\begin{equation}
\label{q4.2}
\begin{aligned}
&P(f_{1, 2}=1|f_{1, 1}=1, T_{f}=3, T_{r}=5)\\
=&\lim_{N\to \infty}\frac{\boldsymbol o_{1, 1}^T\boldsymbol o_{1, 2}}{\boldsymbol o_{1, 1}^T\boldsymbol 1}\bigg |_{T=T_{f}=3}\\
=&\lim_{N\to \infty}\frac{\sum_{t=1}^{4}\boldsymbol o_{1, t}^T\boldsymbol o_{1, t+1}+\boldsymbol o_{1, 5}^TS(\boldsymbol o_{1, 1})}{\left(\sum_{t=1}^{5}\boldsymbol o_{1, t}^T\right)\boldsymbol 1}\bigg |_{T=T_{r}=5},
\end{aligned}
\end{equation}
\begin{equation}
\label{q4.3}
\begin{aligned}
&P(f_{1, 2}=1|f_{1, 1}=1, T_{f}=4, T_{r}=5)\\
=&\lim_{N\to \infty}\frac{\boldsymbol o_{1, 1}^T\boldsymbol o_{1, 2}}{\boldsymbol o_{1, 1}^T\boldsymbol 1}\bigg |_{T=T_{f}=4}\\
=&\lim_{N\to \infty}\frac{\sum_{t=1}^{4}\boldsymbol o_{1, t}^T\boldsymbol o_{1, t+1}+\boldsymbol o_{1, 5}^TS(\boldsymbol o_{1, 1})}{\left(\sum_{t=1}^{5}\boldsymbol o_{1, t}^T\right)\boldsymbol 1}\bigg |_{T=T_{r}=5},
\end{aligned}
\end{equation}
\begin{equation}
\label{q4.3.1}
\begin{aligned}
&P(f_{1, 2}=1|f_{1, 1}=1, T_{f}=T_{r}=5)~~~~~~~~~~~~~~\\
=&\lim_{N\to \infty}\frac{\boldsymbol o_{1, 1}^T\boldsymbol o_{1, 2}}{\boldsymbol o_{1, 1}^T\boldsymbol 1}\bigg |_{T=T_{f}=5}.\\
\end{aligned}
\end{equation}

From (\ref{q4.1})-(\ref{q4.3.1}), it is easy to see that $P(f_{1, 2}=1|f_{1, 1}=1, T_{f}=2, T_{r}=5)=P(f_{1, 2}=1|f_{1, 1}=1, T_{f}=3, T_{r}=5)=P(f_{1, 2}=1|f_{1, 1}=1, T_{f}=4, T_{r}=5)\ne P(f_{1, 2}=1|f_{1, 1}=1, T_{f}=T_{r}=5)$. By a similar approach, we  show that such a relationship holds for general   $T_f$ and $T_r$ (where $T_r\in \mathbb Z_{o}$) as follows.  
 \begin{itemize}
\item When $T=T_f$, there exist  $\boldsymbol o_{1,1}|_{T=T_f}=[o_{1,1}, o_{1, 1+T_f}, o_{1, 1+2T_f},\cdots]^T$, $\boldsymbol o_{1, 2}|_{T=T_f}=[o_{1, 2}, o_{1, 2+T_f}, o_{1, 2+2T_f},\cdots]^T$, $\cdots$, and  $\boldsymbol o_{1, T_f}|_{T=T_f}=[o_{1, T_f}, o_{1, 2T_f}, o_{1, 3T_f},\cdots]^T$.
\item When $T=T_r$, there exist  $\boldsymbol o_{1,1}|_{T=T_r}=[o_{1,1}, o_{1, 1+T_r}, o_{1, 1+2T_r},\cdots]^T$, $\boldsymbol o_{1, 2}|_{T=T_r}=[o_{1, 2}, o_{1, 2+T_r}, o_{1, 2+2T_r},\cdots]^T$, $\cdots$, and  $\boldsymbol o_{1, T_r}|_{T=T_r}=[o_{1, T_r}, o_{1, 2T_r}, o_{1, 3T_r},\cdots]^T$.
\end{itemize}
Accordingly, we have
\begin{equation}
\label{q4.1.1}
\begin{aligned}
&\boldsymbol o_{1, t_f}[n]\boldsymbol o_{1, t_f+1}[n]\\
=&o_{1, t_f+(n-1)T_f}o_{1, t_f+1+(n-1)T_f}\\
=&o_{1, t_r+(l-1)T_r}o_{1,  t_r+1+(l-1)T_r}\\
=&\begin{cases}\boldsymbol o_{1, t_r}[l]\boldsymbol o_{1, t_r+1}[l], \text{when}~ t_r\ne 0\\
\boldsymbol o_{1, T_r}[l]S(\boldsymbol o_{1, 1})[l], \text{when} ~t_r= 0\\
\end{cases}
\end{aligned},
\end{equation}
for the numerator, where $n$ and $l$ are indices, $t_r$ and $t_f$ respectively denote  $t|_{T=T_r}$ and $t|_{T=T_f}$,    and
\begin{equation}
\label{q4.1.2}
\begin{aligned}
t_r=[t_f+(n-1)T_f]\backslash T_r
\end{aligned},
\end{equation}
\begin{equation}
\label{q4.1.3}
\begin{aligned}
l=\left\lfloor\frac{t_f+(n-1)T_f}{T_r}\right\rfloor+1
\end{aligned}.
\end{equation}
On the other hand, for the denominator, there exists $\boldsymbol o_{1, t_f}[n]=o_{1, t_f+(n-1)T_f}=o_{1, t_r+(l-1)T_r}=\boldsymbol o_{1, t_r}[l]$.

Given $T_r\in \mathbb Z_{o}$,  the modulo operation in (\ref{q4.1.2}) leads to $t_r\in\{0, 1, 2, \cdots, T_r-1\}$ when $n$ increases.  Hence, we could obtain
\begin{equation}
\label{q4.1.5}
\begin{aligned}
&P(f_{1, t+1}=1|f_{1, t}=1, T_{f}, T_{r})\\
=&\lim_{N\to \infty} \frac{\boldsymbol o_{1, t}^T\boldsymbol o_{1, t+1}}{\boldsymbol o_{1, t}^T\boldsymbol 1}\bigg |_{T=T_{f}}\\
=&\lim_{N\to \infty}\frac{\sum_{t=1}^{T_r-1}\boldsymbol o_{1, t}^T\boldsymbol o_{1, t+1}+\boldsymbol o_{1, T_r}^TS(\boldsymbol o_{1, 1})}{\left(\sum_{t=1}^{T_r}\boldsymbol o_{1, t}^T\right)\boldsymbol 1}\bigg |_{T=T_{r}},
\end{aligned}
\end{equation}
and then could extend it to the following result,
\begin{equation}
\label{q4.1.5}
\begin{aligned}
&P(f_{i, t+1}=1|f_{i, t}=1, T_{f}, T_{r})\\
=&\lim_{N\to \infty} \frac{\boldsymbol o_{i, t}^T\boldsymbol o_{i, t+1}}{\boldsymbol o_{i, t}^T\boldsymbol 1}\bigg |_{T=T_{f}}\\
=&\lim_{N\to \infty}\frac{\sum_{t=1}^{T_r-1}\boldsymbol o_{i, t}^T\boldsymbol o_{i, t+1}+\boldsymbol o_{i, T_r}^TS(\boldsymbol o_{i, 1})}{\left(\sum_{t=1}^{T_r}\boldsymbol o_{i, t}^T\right)\boldsymbol 1}\bigg |_{T=T_{r}}.
\end{aligned}
\end{equation}

When jointly reviewing the first and second equations  in (\ref{q4.1.5}), it is obvious that $\forall T_f \in\mathbb{T}_f$, $P(f_{i, t+1}=1|f_{i, t}=1, T_{f}, T_{r})$ keeps constant. In other words, $\forall T_f, T_f' \in\mathbb{T}_f$, $P(f_{i, t+1}=1|f_{i, t}=1, T_{f}, T_{r})$ and $P(f_{i, t+1}=1|f_{i, t}=1, T_{f}', T_{r})$ have the same value. Hence, \textit{P1.1}) is proved. On the other hand, we also could see that $P(f_{i, t+1}=1|f_{i, t}=1, T=T_{f})$ and $P(f_{i, t+1}=1|f_{i, t}=1, T=T_{r})$ are different.
Consequently, \textit{Proposition \ref{tt2}}  is proved. Next, we prove \textit{Proposition \ref{tt3}}.

When $T_{r}=6$, we have
\begin{equation}
\label{q4.4}
\begin{aligned}
&P(f_{1, 2}=1|f_{1, 1}=1, T_{f}=2, T_{r}=6)\\
=&\lim_{N\to \infty}\frac{\boldsymbol o_{1, 1}^T\boldsymbol o_{1, 2}}{\boldsymbol o_{1, 1}^T\boldsymbol 1}\bigg |_{T=T_{f}=2}\\
=&\lim_{N\to \infty}\frac{\sum_{t\in\{1, 3, 5\}}\boldsymbol o_{1, t}^T\boldsymbol o_{1, t+1}}{\left(\sum_{t=\{1, 3, 5 \}}\boldsymbol o_{1, t}^T\right)\boldsymbol 1}\bigg |_{T=T_{r}=6},
\end{aligned}
\end{equation}
\begin{equation}
\label{q4.5}
\begin{aligned}
~~&P(f_{1, 2}=1|f_{1, 1}=1, T_{f}=3, T_{r}=6)\\
=&\lim_{N\to \infty}\frac{\boldsymbol o_{1, 1}^T\boldsymbol o_{1, 2}}{\boldsymbol o_{1, 1}^T\boldsymbol 1}\bigg |_{T=T_{f}=3}\\
=&\lim_{N\to \infty}\frac{\sum_{t=1}^{5}\boldsymbol o_{1, t}^T\boldsymbol o_{1, t+1}+\boldsymbol o_{1, 6}^TS(\boldsymbol o_{1, 1})}{\left(\sum_{t=1}^{6}\boldsymbol o_{1, t}^T\right)\boldsymbol 1}\bigg |_{T=T_{r}=6},
\end{aligned}
\end{equation}
\begin{equation}
\label{q4.6}
\begin{aligned}
&P(f_{1, 2}=1|f_{1, 1}=1, T_{f}=4, T_{r}=6)~~~~~~~~~~~\\
=&\lim_{N\to \infty}\frac{\boldsymbol o_{1, 1}^T\boldsymbol o_{1, 2}}{\boldsymbol o_{1, 1}^T\boldsymbol 1}\bigg |_{T=T_{f}=4}\\
=&\lim_{N\to \infty}\frac{\sum_{t\in\{1, 3, 5\}}\boldsymbol o_{1, t}^T\boldsymbol o_{1, t+1}}{\left(\sum_{t=\{1, 3, 5 \}}\boldsymbol o_{1, t}^T\right)\boldsymbol 1}\bigg |_{T=T_{r}=6},
\end{aligned}
\end{equation}
\begin{equation}
\label{q4.7}
\begin{aligned}
&P(f_{1, 2}=1|f_{1, 1}=1, T_{f}=5, T_{r}=6)\\
=&\lim_{N\to \infty}\frac{\boldsymbol o_{1, 1}^T\boldsymbol o_{1, 2}}{\boldsymbol o_{1, 1}^T\boldsymbol 1}\bigg |_{T=T_{f}=5}\\
=&\lim_{N\to \infty}\frac{\sum_{t=1}^{5}\boldsymbol o_{1, t}^T\boldsymbol o_{1, t+1}+\boldsymbol o_{1, 6}^TS(\boldsymbol o_{1, 1})}{\left(\sum_{t=1}^{6}\boldsymbol o_{1, t}^T\right)\boldsymbol 1}\bigg |_{T=T_{r}=6}.
\end{aligned}
\end{equation}

From (\ref{q4.4})-(\ref{q4.7}), it is obvious that $P(f_{1, 2}=1|f_{1, 1}=1, T_{f}=2, T_{r}=6)=P(f_{1, 2}=1|f_{1, 1}=1, T_{f}=4, T_{r}=6)\ne P(f_{1, 2}=1|f_{1, 1}=1, T_{f}=3, T_{r}=6)=P(f_{1, 2}=1|f_{1, 1}=1, T_{f}=5, T_{r}=6)\ne P(f_{1, 2}=1|f_{1, 1}=1, T_{f}=T_{r}=6)$. By a similar approach, we  show that such  a relationship holds for general   $T_f$ and $T_r$ (where $T_r\in \mathbb Z_{e}$) as follows.

When  $T_r$, $T_f\in \mathbb Z_{e}$ and $t_f\in \mathbb Z_{o}$,  the modulo operation in (\ref{q4.1.2}) leads to $t_r\in\{1, 3, \cdots, T_r-1\}$; when  $T_r$,  $T_f\in \mathbb Z_{e}$ and $t_f\in \mathbb Z_{e}$,  the modulo operation in (\ref{q4.1.2}) leads to $t_r\in\{2, 4, \cdots, T_r\}$. Hence, we have
\begin{equation}
\label{q4.8}
\begin{aligned}
&P(f_{i, t+1}=1|f_{i, t}=1, T_{f}, T_{r})\\
=&\lim_{N\to \infty}\frac{\boldsymbol o_{i, t}^T\boldsymbol o_{i, t+1}}{\boldsymbol o_{i, t}^T\boldsymbol 1}\bigg |_{T=T_{f}}\\
=&\begin{cases}\lim_{N\to \infty}\frac{\sum_{t\in\{1, 3, \cdots, T_r-1\}}\boldsymbol o_{i, t}^T\boldsymbol o_{i, t+1}}{\left(\sum_{t\in\{1, 3, \cdots, T_r-1\}}\boldsymbol o_{i, t}^T\right)\boldsymbol 1}\bigg |_{T=T_{r}}, \text{when} ~t|_{T=T_f} \in\mathbb Z_o\\
\lim_{N\to \infty}\frac{\sum_{t\in\{2, 4, \cdots, T_r\}}\boldsymbol o_{i, t}^T\boldsymbol o_{i, t+1}}{\left(\sum_{t\in\{2, 4, \cdots, T_r\}}\boldsymbol o_{i, t}^T\right)\boldsymbol 1}\bigg |_{T=T_{r}}, \text{when} ~t|_{T=T_f} \in\mathbb Z_e
\end{cases}.
\end{aligned}
\end{equation}
On the other hand, when  $T_r\in \mathbb Z_{e}$ and $T_f\in \mathbb Z_{o}$,  the modulo operation in (\ref{q4.1.2}) leads to the same results as (\ref{q4.1.5}).
Thus, by a similar approach as  the proof in the first paragraph below (\ref{q4.1.5}), \textit{Proposition \ref{tt3}}  is proved. 
\end{IEEEproof}
\textit{Proposition \ref{tt2}} and  \textit{Proposition \ref{tt3}} show that there are some regular patterns  as the value of $T$ changes. In addition, such patterns  are different when  the true period is odd or even. Based on the derived two propositions, we have the following two corollaries for the CPbD.

\begin{corollary}
\label{c1}
When $T_{f},T_{f}'\in\mathbb{T}_f$,  $ T_{r}\in \mathbb{T}_r\cap\mathbb Z_{o}$,  there exist:
\begin{description}
\item[\textit{C1.1})]  $D(f_{i, t+1}; f_{i, t})|_{T=T_{f}}\ne D(f_{i, t+1}; f_{i, t})|_{T=T_{r}}$.
\item[\textit{C1.2})]  $D(f_{i, t+1}; f_{i, t})|_{T=T_{f}}= D(f_{i, t+1}; f_{i, t})|_{T=T_{f}'}$.
\end{description}
\end{corollary}
\begin{corollary}
\label{c2}
When $T_{f},T_{f}'\in\mathbb{T}_f$,  $ T_{r}\in \mathbb{T}_r\cap\mathbb Z_{e}$,  there exist:
\begin{description}
\item[\textit{C2.1})]   $D(f_{i, t+1}; f_{i, t})|_{T=T_{f}}\ne D(f_{i, t+1}; f_{i, t})|_{T=T_{r}}$.
\item[\textit{C2.2})]   $D(f_{i, t+1}; f_{i, t})|_{T=T_{f}}= D(f_{i, t+1}; f_{i, t})|_{T=T_{f}'}$ under the condition that  $T_{f}$ and $T_{f}'$ are odd.
\item[\textit{C2.3})]   $D(f_{i, t+1}; f_{i, t})|_{T=T_{f}}= D(f_{i, t+1}; f_{i, t})|_{T=T_{f}'}$ under the condition that  $T_{f}$ and $T_{f}'$ are  even.
\item[\textit{C2.4})]   $D(f_{i, t+1}; f_{i, t})|_{T=T_{f}}\ne D(f_{i, t+1}; f_{i, t})|_{T=T_{f}'}$ under the condition that  $T_{f}$ and $T_{f}'$ are  even and odd respectively.
\end{description}
\end{corollary}
\begin{IEEEproof}
 \textit{Corollary \ref{c1}} and \textit{Corollary \ref{c2}} could be  proved by jointly  applying (\ref{q3.1.2.3.5}), \textit{Proposition \ref{tt2}}, and \textit{Proposition \ref{tt3}}. The detailed proof is omitted here.
\end{IEEEproof}

 \textit{Corollary \ref{c1}} and \textit{Corollary \ref{c2}} show that the CPbD of a CBN given  $T_{r}$  is different from that given  $T_{f}$. But the CPbD has the same value  for all $T_f\in \mathbb{T}_f$, and has the same value  for all $T_r\in \mathbb{T}_r$.  Hence, the CPbD of a CBN is periodical when $T$ monotone increasing. 
 Based on such results,   we propose to use the first peak value of the fast Fourier transformation (FFT) of  the averaged CPbD between $f_{1:M, t}$ and $f_{1:M, t+T_s}$ to find  the empirical $T_p$. As well-known,  the Fourier analysis  methodology is a  feasible approach to find the periodical information in  broad applications. Let nextpow$2(x)$ be a function that  returns  $y$, which is the maximal integer satisfying  $x-2^y\ge0$, and fft($x$) denote the FFT of $x$. The proposed algorithm denoted by $\mathcal T_p$ is given in Table \ref{ta2}.

 \begin{table}[h]
\centering
 \caption{Pseudo-code of  $\mathcal{T}_p$.}
  \label{ta2}
\begin{tabular}{ c | l }
  \hline 
  1) &  $l=$nextpow$2(T_s^*)$; \\
  2) & for $T_s=2:1:2^l+1$ \\
   &~~~for $t=1:1:T_s$ \\
   &~~~~~~$\boldsymbol D_p[t]= \boldsymbol D_{t, t+T_s}$;\\
   &~~~end\\
   &~~~$\boldsymbol D[T_s-1]=\frac{||\boldsymbol D_p||_1}{T_s}$;\\  
   &end\\
3) & $\tilde{\boldsymbol D}= \text{fft}(\boldsymbol D)$;\\
 4) & for $u=3:1:2^{l-1}-1$ \\
   &~~~if $\tilde{\boldsymbol D}[u-1]>\tilde{\boldsymbol D}[u-2]$ and $\tilde{\boldsymbol D}[u-1] > \tilde{\boldsymbol D}[u]$,\\
   &~~~then break and $T_p=\max_{n\in\mathbb Z}\frac{1}{n(u-1)}$, st. $\frac{1}{n(u-1)}<T_s^*$;\\
   &end\\
5) &if $u=2^{l-1}+1$,\\
   &then set $l=l+1$ and go to 2).\\
  \hline  
\end{tabular}
   \end{table}
  \subsection{Overall CBN Learning Scheme}
Based on  (\ref{2.a.88}) and the algorithms $\mathcal T_s$ and $\mathcal T_p$, the complete  learning scheme for the CBN is given in Table \ref{t3}, by which we could efficiently  obtain the CBN with measured edges and the related conditional probability table.
 \begin{table}[h]
\centering
 \caption{Pseudo-code of  the proposed learning scheme}
  \label{t3}
\begin{tabular}{ c | l }
  \hline 
  1) & for $m=1:1:M$ \\
      &~~~$\boldsymbol T_s[m]=\mathcal T_s|m$;\\
      & end \\
	& $T_s=\max_m \boldsymbol T_s[m]$;\\
  2) & set $l=T_s$ in 1) of $\mathcal T_s$ and perform  $\mathcal T_s$ to obtain $T_s^*$; \\
  3) & perform  $\mathcal T_p$ with $T_s^*$ to obtain $T_p$; \\
  4) & obtain $T$ by using (\ref{q3.1.2.3.19.2})-(\ref{q3.1.2.3.22}) with $T_s^*$ and $T_p$;\\
  5) & obtain $\boldsymbol B_{1:T-1}$ and $\bar{\boldsymbol D}_{1:T-1}$ with $T$.\\
  \hline  
\end{tabular}
   \end{table}

\section{Simulation Results}
In this section, we first give the comparison of computational  complexity between the proposed CPbD algorithm and the conventional one \cite{Cheng:2002:LBN:570380.570382}. Then we present the learning outcomes. 

\subsection{Comparison of Computational Complexity}
\begin{figure}[t]
  \centering
\includegraphics[width=7.8cm]{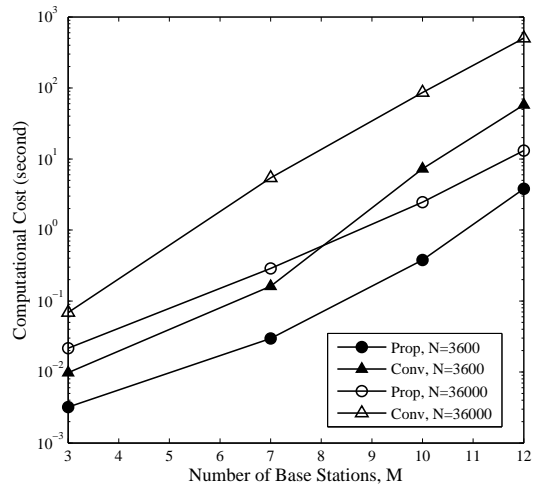}
  \caption{Comparison of computational cost between ``Prop" and ``Conv". ``Prop" and ``Conv" are abbreviations  for our proposed scheme and the conventional one proposed in \cite{Cheng:2002:LBN:570380.570382}, respectively.}
 \label{f6.1}
\end{figure}

\begin{figure}[t]
  \centering
\includegraphics[width=7.8cm]{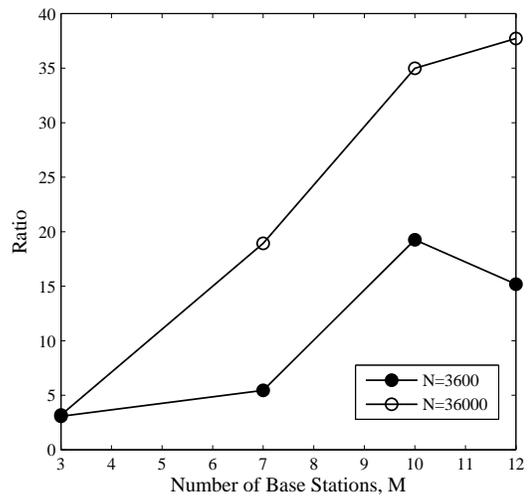}
  \caption{Ratio of the Conv's computational cost  to the  Prop's.}
 \label{f6.2}
\end{figure}

The computational complexity is evaluated by running the Matlab codes of the proposed algorithm and the conventional one in the same desktop. The corresponding  running time is recorded to reflect the real computational cost. In this subsection,  the learning complexity is evaluated when considering the CBN structure that   only consists of one C-clique,  since the learning time of the whole  CBN structure   is linear to that of a C-clique. 
Additionally,  the impacts of observation numbers and variable  numbers  are considered  for a comprehensive  comparison. In our simulations, the observations are generated based on $M$ binomial distributions \footnote{Actually, the distributions of  observations do not impact  the computational cost of learning.}. The simulated results are presented in Fig. \ref{f6.1} and Fig. \ref{f6.2}, where the computational cost is the total running time for obtaining the conditional probability table and the measurement of every edge. It is obvious that the computational cost of the proposed algorithm is much  less  than the conventional one, especially as the numbers of base stations and observations increase. In particular, when $M=12$ and $N=36000$, the conventional algorithm demands for $502.2$ seconds to learn. It implies that  the conventional algorithm is not suitable for online learning even when the number of  variables is not large.
In Fig. \ref{f6.2}, we draw the ratio between the ``Conv" and ``Prop" costs over $M$, where  it is obvious that the proposed scheme is more efficient. 
 According to   Fig. \ref{f6.2}, the proposed algorithm only requires  $1/38$ of the time cost as the conventional one when  $M=12$ and $N=36000$.  In addition, note that the proposed algorithm could adapt to any number of variables, while the conventional one cannot.

 \begin{figure*}[t]
  \centering
\includegraphics[width=11cm]{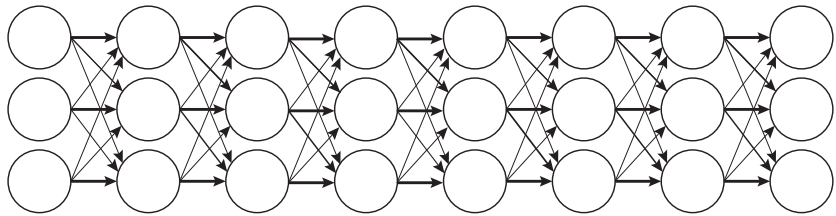}
  \caption{Illustration of learned CBN. The width of edges reflects the magnitude of dependence.}
 \label{f6.3}
\end{figure*}

\subsection{Learning Outcomes}
The simulation scenario is illustrated by Fig. \ref{f1}, where mobile users move from the right to the left and access one of the three primary base stations according to their locations. The base stations are respectively labeled $\#1$, $\#2$, and $\#3$ from right to left.  The simulation configuration is as follows: The arrival of mobile users follows a Poisson distribution with mean $\lambda_s=1$;    the length of the road is $600$ meters, covered by the three  cells; the data traffic arrival rate  at each user follows a Poisson distribution with mean $\lambda_t=0.002$;\footnote{The value of  $\lambda_t$ is set small since we have to ensure that the primary base stations have some idle slots.} the  service time of data traffic follows an exponential distribution with a mean of $2$ seconds; and spectrum sensing is performed  once  per second.

In this paper, we are interested in the relationship among base station activities, which is introduced by the user movement. Such statistical correlation is useful for CR operation but has not been well studied in the past. To show the learning results, the data of observation is collected in three different cases corresponding to  the uniformly distributed user speed regions as   $[43.2~ 72]$, $[72~129.6]$, and $[100.8~158.4]$ kilometers per hour, respectively. The learning results of $T$ (empirical period of CBN structure) are presented in Table \ref{t5}, where the unit of $T$ is second. It is obvious that the period $T$ is similar under different numbers of observations, which shows that the proposed algorithm could work robustly. According to the learning results, it could be concluded that the statistical relationship of the observed three base stations are mainly affected   by the user speed, and the period is inversely proportional to the user speed. 
 Since the case of $N=3600\times 10$ has  the most number of observations, it could show the most trustable  results when compared  with the other two.  Interestingly, we observe that $T\times v\approx\frac{600}{2}$ from our simulation results, where $v$ is the averaged speed over the speed region. Actually, such relationship is reasonable, since  $\frac{600}{2}$ is the mean of the distances between the  mobile users and the left end of the simulated road.
 \begin{table}[h]
\centering
 \caption{Learning Results of $T$.}
  \label{t5}
\begin{tabular}{| c | c| c | c | }
  \hline 
  Speed Region &  $N=3600$ & $N=3600\times 5$ & $N=3600\times 10$\\
\hline
[43.2~ 72] &  $T=16$ & $T=16$ & $T=18$\\
\hline
[72~129.6] &  $T=10$ & $T=12$ & $T=12$\\
\hline
[100.8~158.4] &  $T=8$ & $T=10$ & $T=8$\\
  \hline  
\end{tabular}
   \end{table}

Next, we show the learning results of CPbD. For clarity of expression, we only show the results in the case of $[100.8~158.4]$ and $N=3600\times 10$. When $T=8$, the learned statistical pattern $\bar{\boldsymbol D}_{1:7}$ is given as follows,
\begin{align}
\bar{\boldsymbol D}_1=\begin{vmatrix} 
     1.0000  &  0.3818  &  0.0917\\
    0.0844  &  1.0000  &  0.3592\\
    0.2058  &  0.1779  &  1.0000
 \end{vmatrix},\\
\bar{\boldsymbol D}_2=\begin{vmatrix} 
    1.0000  &  0.5724  &  0.2733\\
    0.3201  &  1.0000  &  0.4926\\
    0.2173  &  0.2721  &  1.0000
 \end{vmatrix},
\\
\bar{\boldsymbol D}_3=\begin{vmatrix} 
    1.0000  &  0.5041  &  0.1137\\
    0.3084  &  1.0000  &  0.5247\\
    0.2524  &  0.2279  &  1.0000
 \end{vmatrix},
\\
\bar{\boldsymbol D}_4=\begin{vmatrix} 
    1.0000  &  0.4337  &  0.0345\\
    0.0527  &  1.0000  &  0.3740\\
    0.2376  &  0.2636  &  1.0000
 \end{vmatrix},
\\
\bar{\boldsymbol D}_5=\begin{vmatrix} 
    1.0000  &  0.4379  &  0.1431\\
    0.2354  &  1.0000  &  0.6035\\
    0.0716  &  0.0659  &  1.0000
 \end{vmatrix},
\\
\bar{\boldsymbol D}_6=\begin{vmatrix} 
    1.0000  &  0.5247  &  0.3372\\
    0.2607  &  1.0000  &  0.5282\\
    0.2976  &  0.2644  &  1.0000
 \end{vmatrix},
\\
 \bar{\boldsymbol D}_7=\begin{vmatrix} 
    1.0000  &  0.5490  &  0.2898\\
    0.2796  &  1.0000  &  0.5051\\
    0.0760  &  0.1120  &  1.0000
 \end{vmatrix}.
\end{align}

Apparently, the entries in the upper triangular part of $\bar{\boldsymbol D}$ is generally larger than the entries in the lower triangular part, which implies that the status of base stations $\#2$  and $\#3$  is heavily impacted by  base station $\#1$, but the reverse does not hold. In addition, we use the learned CPbD $\bar{\boldsymbol D}_{1:7}$ to draw a CBN shown in Fig. \ref{f6.3}, where the line width is proportional to the magnitude of dependence and   we  see that the trend of dependence between  base station $\#1$ and base station $\#3$ increases as $t$ increases  during a period. It is due to the fact that  the users served by  base station $\#1$ at the beginning of a period will arrive in the range of base station $\#3$ at the end of the period. 
 Note that if we add a binary  thresholding to make decisions on the existence of edges, which is used for binary testing of each edge in  \cite{Cheng:2002:LBN:570380.570382}, the learning outcomes will be similar to those  in \cite{Cheng:2002:LBN:570380.570382}. The estimated conditional probability table $\boldsymbol B_{1:7}$ is also reported in (\ref{f56})-(\ref{f62}). Based on the learned conditional probability table, we could  generate the joint distribution of the on/off behaviors of the three primary base stations and  predict the status of network for  the future, to serve broader CR applications. 
 
In brief, the simulated outcomes show that the proposed   CBN  structure learning algorithm  not only can quantify the strength  of dependence  but also correctly reflect the  unidirectional statistical pattern caused by the mobile user's one-way movement, indicating that   our learning scheme can efficiently learn the CBN structure to represent the true   statistical behavior of the underlying network.

\section{Conclusion}
In this paper, we proposed a learning scheme to obtain the statistical  pattern of a primary network's activity  in both spatial and  temporal domains simultaneously.
 The proposed scheme incurs significantly  lower computational complexity  when compared  with the traditional ones. Additionally,  it is capable of  learning the statistical period of  the CBN structure. By simulations, we show that the learning results also correctly reflect certain network behavior beyond spectrum usage, which could be useful for broader CR network control applications.

\begin{align}
\label{f56}
\boldsymbol B_1=\begin{vmatrix} 
    0.6664  &  0.6664  &  0.6664\\
    0.7953   &  0.7342  &  0.0010\\
    0.7403  &  0.1119  &  0.8881\\
    0.7649  &  0.0139  &  0.0139\\
    0.1674  &  0.7329  &  0.7329\\
    0.1995  &  0.7495  &  0.0208\\
    0.0176  &  0.2048  &  0.7495\\
    0.0115  &  0.0109  &  0.0232\\
 \end{vmatrix},
\\
\boldsymbol B_2=\begin{vmatrix} 
    0.9990 &   0.9990  &  0.9990\\
    0.8688   & 0.8471   & 0.0227\\
    0.7994   & 0.2006  &  0.6996\\
    0.7438   & 0.0187   & 0.0212\\
    0.2672   & 0.7661   & 0.8992\\
    0.2024   & 0.7266   & 0.0128\\
    0.0089   & 0.1778   & 0.7790\\
    0.0142   & 0.0118   & 0.0177\\
 \end{vmatrix},
 \\
\boldsymbol B_3=\begin{vmatrix} 
    0.5001 &   0.9990  &  0.9990\\
    0.7994  &  0.8357  &  0.0192\\
    0.7935  &  0.2652  &  0.7935\\
    0.7687  &  0.0164  &  0.0215\\
    0.2075  &  0.7925  &  0.7581\\
    0.2000  &  0.7558  &  0.0136\\
    0.0090  &  0.1919  &  0.7326\\
    0.0112  &  0.0124  &  0.0170\\
 \end{vmatrix},
  \\
\boldsymbol B_4=\begin{vmatrix} 
    0.7994  &  0.7994 &   0.9990\\
    0.7772  &  0.7957  &  0.0195\\
    0.7415  &  0.1942  &  0.7415\\
    0.7553  &  0.0138  &  0.0241\\
    0.2339  &  0.7329  &  0.7994\\
    0.2252  &  0.7533  &  0.0286\\
    0.0094  &  0.2181  &  0.7359\\
    0.0126  &  0.0126  &  0.0228\\
 \end{vmatrix},
   \\
\boldsymbol B_5=\begin{vmatrix} 
    0.5001 &   0.9990 &   0.7496\\
    0.7772  &  0.7033  &  0.0380\\
    0.8372  &  0.2438  &  0.7023\\
    0.7815  &  0.0138  &  0.0215\\
    0.1436  &  0.8921  &  0.7139\\
    0.2030  &  0.7397  &  0.0161\\
    0.0168  &  0.2140  &  0.7505\\
    0.0136  &  0.0139  &  0.0204\\
 \end{vmatrix},
    \\
\boldsymbol B_6=\begin{vmatrix} 
    0.9990  &  0.9990  &  0.9990\\
    0.8254  &  0.7387  &  0.0227\\
    0.6919  &  0.1290  &  0.8199\\
    0.7568  &  0.0131  &  0.0083\\
    0.1570  &  0.8431  &  0.6559\\
    0.1948  &  0.7329  &  0.0241\\
    0.0372  &  0.2062  &  0.7173\\
    0.0113  &  0.0137  &  0.0196\\
 \end{vmatrix},
     \\
\boldsymbol B_7=\begin{vmatrix} 
    0.9990  &  0.9990  &  0.5002\\
    0.7495  &  0.7703   & 0.0218\\
    0.7709  &  0.1436  &  0.7994\\
    0.7599  &  0.0132  &  0.0156\\
    0.2313  &  0.7687  &  0.6536\\
    0.1843  &  0.7866  &  0.0214\\
    0.0092  &  0.2055  &  0.7250\\
    0.0119  &  0.0128  &  0.0219\\
 \end{vmatrix}.
 \label{f62}
\end{align}

%
%

\end{document}